\documentclass[journal]{IEEEtran}  

\IEEEoverridecommandlockouts                              

\usepackage{siunitx}
\usepackage{graphicx}
\usepackage{caption}
\usepackage{subcaption}
\usepackage{array,multirow}
\usepackage{float}
\usepackage{hyperref}
\usepackage{url}
\usepackage{tabularx}
\usepackage{adjustbox}
\usepackage[dvipsnames]{xcolor}
\usepackage{color}
\hypersetup{
    colorlinks=true,
    linkcolor=blue,
    citecolor=blue,      
    urlcolor=blue,
}
\usepackage{mathptmx} 
\usepackage{amsmath} 
\usepackage{soul}

\title{\LARGE \bf
An Uncertainty-Aware Deep Learning Framework for Defect Detection in Casting Products
}

\author{Maryam Habibpour, Hassan Gharoun, AmirReza Tajally, Afshar Shamsi, Hamzeh Asgharnezhad\\ Abbas Khosravi ~\IEEEmembership{Senior,~IEEE,} and Saeid Nahavandi~\IEEEmembership{Fellow,~IEEE,}

\thanks{This  research  was partially  supported  by  the  Australian  Research  Council’s  Discovery  Projects funding scheme (project DP190102181).}
\thanks{M. Habibpour is with Lian Sazeh group, Tehran, Iran (e-mail: E.maryamhabibpour@gmail.com)}
\thanks{H. Gharoun is with the School of Industrial Engineering, College of Engineering, University of Tehran, Tehran, Iran (e-mail: h.gharoun@alumni.ut.ac.ir)} 
\thanks{A. Tajally is with the Department of Industrial Engineering, university of Tehran, Tehran, Iran (e-mail: Amirreza73tajally@gmail.com)}
\thanks{A. Shamsi, and H. Asgharnezhad are individual researchers, Tehran, Iran (e-mail: \{afshar.shamsi.j, Hamzeh.asgharnezhad\}@gmail.com)}
\thanks{A. Khosravi,  and S. Nahavandi are with the Institute for Intelligent Systems Research and Innovation (IISRI), Deakin University, Australia (e-mail: \{abbas.khosravi, saeid.nahavandi\}@deakin.edu.au)}
}

\begin{document}

\maketitle
\thispagestyle{empty}
\pagestyle{empty}

\begin{abstract}
Defects are unavoidable in casting production owing to the complexity of the casting process. While conventional human-visual inspection of casting products is slow and unproductive in mass productions, an automatic and reliable defect detection not just enhances the quality control process but positively improves productivity. However, casting defect detection is a challenging task due to diversity and variation in defects’ appearance. Convolutional neural networks (CNNs) have been widely applied in both image classification and defect detection tasks. Howbeit, CNNs with frequentist inference require a massive amount of data to train on and still fall short in reporting beneficial estimates of their predictive uncertainty. Accordingly, leveraging the transfer learning paradigm, we first apply four powerful CNN-based models (VGG16, ResNet50, DenseNet121, and InceptionResNetV2) on a small dataset to extract meaningful features. Extracted features are then processed by various machine learning algorithms to perform the classification task. Simulation results demonstrate that linear support vector machine (SVM) and multi-layer perceptron (MLP) show the finest performance in defect detection of casting images. Secondly, to achieve a reliable classification and to measure epistemic uncertainty, we employ an uncertainty quantification (UQ) technique (ensemble of MLP models) using features extracted from four pre-trained CNNs. UQ confusion matrix and uncertainty accuracy metric are also utilized to evaluate the predictive uncertainty estimates. Comprehensive comparisons reveal that UQ method based on VGG16 outperforms others to fetch uncertainty. We believe an uncertainty-aware automatic defect detection solution will reinforce casting productions quality assurance.

\end{abstract}


\section{Introduction and Literature Review}
\label{sec:intro}
\IEEEPARstart{Q}{UALITY} is paramount in manufacturing processes, particularly those involving casting, so quality along with other main objectives should be retained and elevated by manufacturers. Only high-quality products can prevail on the market on a long-term basis. Defects in the casting process can negatively affect the quality of the final product, resulting in massive losses for manufacturing businesses \cite{rajkolhe2014defects}. Extremely high-volume production exacerbates this issue and demands closer inspection. Accurate detection of defects in casting-related industries is a relevant issue that draws the attention of machine learning (ML) and computational intelligence communities, where a wide variety of automatic solutions have been proposed \cite{rajkolhe2014defects},
\cite{zhu2019defect}, \cite{yang2018transfer}.



 More generally, the literature on the detection of defects in casting products contains a large number of methodologies, the majority of them revolve around deep learning (DL) models and, in particular, convolutional neural network (CNN) algorithms. Recently, the resurgence of deep neural networks has achieved significant breakthroughs in the area of defect detection of products.
 
 The success of CNN algorithms with arbitrary depth and non-linearity depends heavily on the abundant amount of properly labeled images used to train the effective deep network, which must efficiently represent a broad spectrum of features. There are limited resources for labeling thousands of images in industrial environments (i.e., experts), which makes it difficult to develop reliable CNNs for defect inspection. Accordingly, leveraging the transfer learning framework, the VVG16 pre-trained model was proposed as a feature extractor to construct a defect detection model; the extracted features were then processed by an improved CNN algorithm and conventional SVM model. The obtained accuracies for a binary class are 97.02\% and 74.20\%, respectively \cite{zhu2019defect}. This procedure was also employed for online training and classification of Mura defects, in which the AlexNet feature extractor was used to extract meaningful features from input images \cite{yang2018transfer}. Authors in \cite{mittel2019vision} applied two pre-trained models to extract features and then transferred them to the CNN model for processing. The experimental results in the paper show that GoogleNet outperforms AlexNet with an accuracy of 0.998 and an F1-score of 0.836. 

All papers in the literature report promising results for CNNs trained using a limited number of images, but they unrealistically assume that any type of sample of the product is available during the model initial training, and ignore the real-world risks that confound computer vision techniques. Vision-based models trained with a limited set of samples without exerting diverse scenarios easily fail in real-world applications. Moreover, the limited number of samples raises concerns regarding epistemic uncertainty \cite{shamsi2021uncertainty}. In practice, input distributions could be shifted from the training distribution for a variety of reasons, such as sample selection bias and non-stationary environments, leading to the lack of robustness of modern CNN models \cite{ovadia2019can}. Discriminative neural networks with numerous pathologies are vulnerable to maliciously manipulated and almost imperceptible perturbation over the original data (to human vision systems), resulting in erroneous predictions over the manipulated samples \cite{akhtar2018threat}. These complex models are pathologically prone to yield  overly-confident predictions, even when their predictions are wrong \cite{nguyen2015deep}. Mentioned reasons raise questions about the performance of neural networks and cause significant security concerns in the practical deployment. In the product quality control process, where every mistake can trigger irreparable damage, the reliability of the predictions should be more important than providing predictions for all query samples. If ML-based systems would reliably recognize cases in which they expect to underperform, and consequently abstain from classification and instead defer to an expert engineer, they could more securely be deployed.

In response, the implementation of an uncertainty quantification (UQ) technique is essential to build uncertainty-aware neural networks. UQ methodology performs distribution estimation accompanied by class prediction, leading to a more informed decision and improving the quality of predictions. This procedure assesses risk accurately and permits a system to refrain from suboptimal decisions due to low confidence. Leveraging the UQ technique, whenever the model encounters an incoming image related to the casting samples as being hitherto unknown, it attempts to communicate with its knowledge to know it,  which leads to a report of uncertainty. Here, engineers' intervention is necessary to prevent mistakes and potential losses. Providing a report of uncertainty regarding generated predictions by these powerful black box predictors is critical to be widely adopted as a trustworthy defect detection tool.

Our contribution in this paper has two parts. First, since providing thousands of casting product images is labor-intensive for developing a dependable model to detect defects, we apply four pre-trained CNN models (VGG16, DenseNet121, InceptinResNetV2, and ResNet50) to extract informative features from casting product images. Extracted features are processed by several ML models to perform the classification task. The obtained results from variant CNN architectures and classifiers are measured by variant performance metrics so that we achieve a general evaluation. Secondly, to quantify the uncertainty of generated predictions and achieve a reliable classification we create an ensemble of deep neural networks trained using variant features. Given the UQ confusion matrix and uncertainty accuracy metric, we then examine the predictive uncertainty estimates. The evaluation of uncertainty is carried out much the same way as a binary classification evaluation.

\textit{Structure of the paper}: The UQ technique is discussed in more detail in section \ref{sec:UQ}. We explain the transfer learning procedure in section \ref{sec:TL}. Description of dataset and details of the experimental setup is provided in section \ref{sec:ExperimentSetup}. Section \ref{sec:discussion_result} presents the discussions and empirical results. Conclusion and future research follow in section \ref{sec:conclusion}.

\section{Using uncertainty}
\label{sec:UQ}
The ability to communicate uncertainty is a sign of the trustworthiness of a model, which can improve user experience and facilitate model adoption. The main considerations presented to quantify uncertainty for DL models encompass two types  \cite{depeweg2019modeling}: first, aleatoric uncertainty (also known as indirect uncertainty) is related to the noise intrinsic or class overlap in input data, which partially reducible by adding extra features; second, epistemic uncertainty (also known as direct uncertainty) is related to the model variability which is quite reducible by adding more data in input regions where the training set was sparse. This study prioritizes the examination of epistemic uncertainty because it relates to models’ generalization power for novel samples \cite{quan2019survey}. The proposed method to quantify uncertainty and metric to evaluate the predictive uncertainty estimates are described below.

\subsection{Non-Bayesian method to quantify Uncertainty}
There is a general acceptance that probabilistic Bayesian models have shown accepted results for measuring uncertainties and reasoning about confidences. However, they are developed with simple modeling assumptions and are computationally expensive. This has led to numerous approximate estimation strategies being developed for neural networks. One of these strategies is ensemble of neural networks, a widely used method for measuring epistemic uncertainties, and maintaining robustness even under dataset shift \cite{ovadia2019can}. Ensemble trains multiple neural networks in various ways to obtain multiple plausible fits. At inference time, the disagreement between ensemble members’ predictions yields model uncertainty. While all networks are expected to behave similarly in areas with enough training data, inconsistent outcomes will arise in areas with sparse training data. The outcomes are subjected to an empirical distribution, which can be exerted to determine the mean value and the confidence measure in terms of the distributional variance. The mean value of one test data point is calculated as follows (\ref{eq:1}):

\begin{equation}
\hat{p}\left(y| x \right)=\frac{1}{M} \sum_{i=1}^{M} p{\theta_i} \left(y | x \right) 
\label{eq:1}
\end{equation}

\noindent where ${\theta_i}$ denotes the set of parameters for $i^{th}$ network.
The entropy metric can be exerted as the uncertainty estimate to determine the distribution variance (\ref{eq:2}).

\begin{equation}
H\left( \hat{p}\left(y| x \right) \right) = \sum_{i=0}^{C} \hat{p} \left({y_i} | x \right) \log \hat{p} \left({y_i} | x \right)
\label{eq:2}
\end{equation}

\noindent Here, $C$ refers to all classes. The low entropy value indicates that all ensemble members make similar predictions.

\subsection{Evaluation of predictive Uncertainty}
\begin{figure}[thpb]
      \centering
      \includegraphics[width=\linewidth]{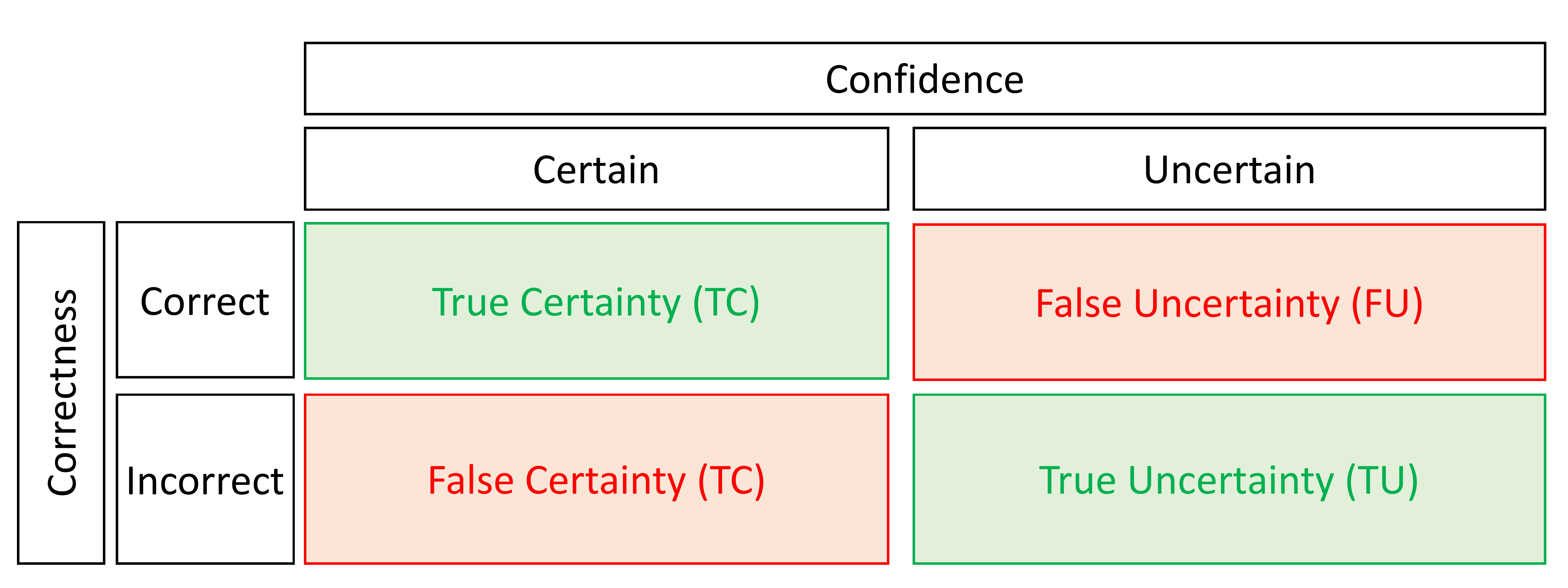}
      \caption{The UQ confusion matrix}
      \label{fig:confmatrix}
   \end{figure}

Since classification tasks are done by conventional supervised learning models, the confusion matrix permits visualization and comparison of their performance. The layout of such a matrix indicates the classifier’s performance on the set of test data. Predicted test data are generally divided into two categories, correct and incorrect. \cite{asgharnezhad2020objective} proposed an idea similar to the confusion matrix to evaluate the predictive uncertainty estimates. In order to do this, a threshold is determined to cast predictions into certain and uncertain categories. The combined results of the two groups are shown in Fig. \ref{fig:confmatrix}. There are four possible combinations, including true certainty (TC) where the prediction is correct and certain, true uncertainty (TU) where the prediction is incorrect and uncertain, false uncertainty (FU) where the prediction is correct and uncertain, false certainty (FC) where the prediction is incorrect and certain. Both TC and TU are ideal results and correspond to TN and TP in the typical confusion matrix respectively. While FU is close to desirable result, FC is the worst-case scenario. Hence, the uncertainty accuracy metric is defined as follows\cite{khaledyan2021confidence} (\ref{eq:3}):

\begin{equation}
uncertainty \ accuracy = \frac{TU + TC}{TU + TC + FU + FC}
\label{eq:3}    
\end{equation}

\noindent In this metric, 1 is the best result and 0 is the worst. The best result implies that the model is aware of what it knows and what it does not know.

\section{Transfer Learning}
\label{sec:TL}
\begin{figure}[!thpb]
      \centering
      \includegraphics[scale=0.085]{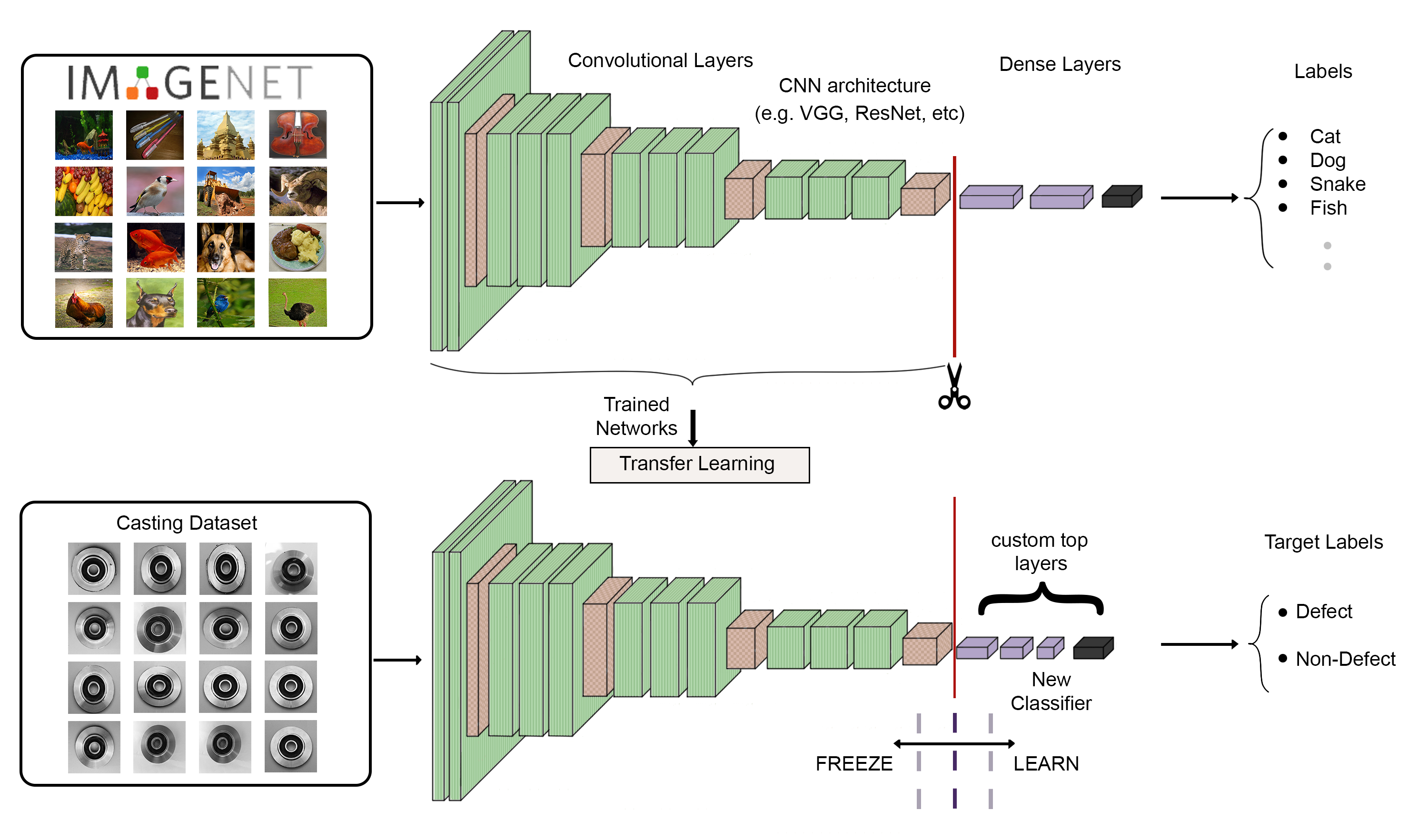}
      \caption{Illustration of the proposed transfer learning framework; using pretrained CNNs on manufacturing casting images for defect detection}
      \label{fig:transferlearning}
   \end{figure}

In this study, concerning the binary classification problem of defect detection, the transfer learning approach is employed. Generally, training a CNN-based deep learning classification model involves two main issues: first, it requires a massive dataset, and second, tuning the model’s parameters is a time-consuming computational process. In the transfer learning approach, a large dataset has trained a pre-trained CNN model, and parameters initiated \cite{rangarajan2020disease}. The in-depth CNN model is prone to overfitting in training with a small dataset, while these pre-trained models with initialized parameters shorten the training process and are robust to overfitting \cite{zhang2021covid}. 
Here, based on the concept of transfer learning, four pre-trained CNN models, particularly VGG16, DenseNet121, ResNet50, and InceptionResNetV2 are employed. These networks, containing millions of parameters to be tuned, have been trained with the ImageNet dataset and can extract necessary features \cite{alam2020automatic}. The ImageNet data set is not applicable for manufacturing casting applications. Here, these pre-trained models trained with the non-casting dataset are employed to apply the gained information from another problem to a casting defect detection problem. In this study, to make the classification problem easier, the feature extraction usefulness of these pre-trained models is examined in manufacturing casting process applications. 
As shown in Fig. \ref{fig:transferlearning}, to fine-tune pre-trained CNNs: first, the last pooling layer is omitted to keep informative features, second, the convolutional layers’ weights are frozen, and last, various machine learning classification algorithms are superseded forepart’s fully connected layers of pre-trained CNNs. These modified pre-trained models are used to extract features of input images. Then, extracted features proceed to machine learning classification algorithms to detect defects. 

\section{Experiment Setup}
\label{sec:ExperimentSetup}
\subsection{Data Set}

The proposed method has been applied to a publicly available dataset of manufacturing casting of submersible pump impeller \cite{dabhi_2020}. This impeller is a rotary iron disc connected to the motor by shaft and rotates at high speed (typically 500-5000rpm) to speeds up the fluid out through the impeller vanes into the pump casing. Occurring an unwanted phenomenon during the iron casting process of this product is known as a defect. 
In the utilized data set, containing 1300 samples of 512×512 grey-scaled image data, two types of images have been used: defect (781 samples) and non-defect (519 samples) images. Before commencing the processing, the whole dataset is divided into 75\%-25\% as training and test sets. 

\subsection{Pre-trained Models}
Four pre-trained CNN models employed in this study to extract features, are briefly presented below. 
\begin{itemize}
\item VGG16: VGG16 is built on the stacked architecture of AlexNet trained on the ImageNet dataset.  Here, the network contains 13 convolution layers with one stride, and padding of 1 stacked with 3 fully connected layers is applied. VGG16 utilizes a filter with a small receptive field of 3×3 and uses total 5 max-pooling layers of 2×2 with stride 2 and no padding \cite{rezende2018malicious}. 
\item ResNet50: ResNet50 stands for residual network containing 50 layers. ResNet uses identity mapping to predict the required delta for reaching the final prediction from one layer to the other. Identity mapping and shortcut path allow ResNet to bypass unnecessary layers to flow gradient in the backpropagation process. This structure enables ResNet to reduce the vanishing gradient problems \cite{theckedath2020detecting}. 
\item DenseNet121: DenseNet consists of dense blocks with various filters between each block.  Unlike traditional neural networks, in which only adjacent layers have connections, each layer receives the union output of all previous layers as the input in dense blocks. Transition layers connect dense blocks and downsample the features by convolution and pooling layers \cite{huang2017densely}. DenseNet reduces exploding or gradient vanishing problems and, with fewer parameters, decreases computational complexity \cite{wang2019captcha}.    
\item InceptionResNetV2: InceptionResNetV2 is built based on combining Inception and ResNet architectures. Inception blocks are a form of multi-branch structure and use various filters with the concatenation in each block \cite{kamble2018automated}. Using residual shortcuts in InceptionResNetV2 simplifies inception blocks and allows training for very deep architecture.  
\end{itemize}
During the transfer learning process, network weights are kept frozen. In this study, the input image size for the InceptionResNetV2 is 299×299 and, for the rest of the models, it is 244×244. Table \ref{tab:CNNs_info} provides general information about proposed pre-trained models. 

\begin{table}[t]
\centering
\small
\caption{Information about four considered architectures for transfer learning}
\label{tab:CNNs_info}
\resizebox{\columnwidth}{!}{%
\begin{tabular}{lccc}
\hline
Architecture & Input Size & Number of Features & Number of Parameters \\
\hline
VGG16 & 244×244 & 25,088 & 14,714,688 \\
ResNet50 & 244×244 & 100,352 & 23,587,712 \\
DenseNet121 & 244×244 & 50,176 & 7,037,504 \\
InceptionResNetV2 & 299×299 & 98,304 & 54,336,736\\
\hline
\end{tabular}%
}
\end{table}

\subsection{Classification method}
In this study, the global approach in the classification task consists of two main steps: the first step focuses on image feature extraction by employing multiple pre-trained CNN models, while the second step emphasizes processing the extracted features by developing various classifiers. In the following, a brief description of the proposed classifiers is presented. 
\begin{itemize}
\item K-nearest neighbors (KNN): KNN is a simple classification method based on instance-based learning. For each test sample, KNN searches for its k nearest training instances in the space of training data features and computes its label based on the majority voting of its neighbors. Generally, KNN performance depends on the choice of k parameter and distance metric. Euclidean distance (\ref{eq:4}) is the most widely used distance metric in the KNN algorithm, especially when there is no or little knowledge on data distribution \cite{weinberger2009distance}. KNN algorithm is highly sensitive to training data set sparseness and distribution form \cite{wang2018comparative}. 
In this study, a KNN with k equals 2, and Euclidean distance is used.

\begin{equation}
d\left(x, y \right)=\left(\sum_{i=1}^{n}\left({x_i},{y_i} \right)^2 \right)^\frac{1}{2}
\label{eq:4}  
\end{equation}

\noindent where $i = 1,…,n$ is feature space

\item Naive Bayes (NB): In pattern recognition, Naive Bayes is a simple probabilistic method using Bayes theorem for calculating the conditional probability of a new instance assigning to each class. This method rests on the assumption of independence between the pair of feature space. Given a new instance $X = {{x_1},...,x{_n }}$ in an n-dimension feature space and predefined classes   $C = {{C_1},...,{C_J }}$, the NB classifier estimates probability as (\ref{eq:5}):

\begin{equation}
P({C_j}| X) =\frac{P(X|{C_j}).P({C_j})}{P(X)} ; 1\leq j\leq J 
\label{eq:5}  
\end{equation}

\noindent where: $P({C_j})$ and $P(X)$ are the probability of class $j$ and instance $x$ independent of each other,
$P({C_j})$ is the conditional probability of class $j$ where instance $X$ belongs to ${C_j}$, and
$P(X|{C_j})$ is the probability of instance $X$ belonging to $C_j$.\\
Since $P(X)$ is the same for all classes, NB classifier assigns new instance X to the class with maximum $P(X|{C_j}).P({C_j})$ \cite{yilmaz2019locally}. In this study, Gaussian Naive Bayes is employed where the feature values follow Gaussian distribution.

\item Gaussian process (GP): GP estimates the probability distribution over possible functions, so that any set of functions has a joint Gaussian distribution. GP model is described by its mean function $(E[f(x)]=\mu(x))$ which is considered 0) and covariance function: it is also named kernel ($Cov[f(x),f(x)]=k(x,x)$). Concerning a binary classification with the given data composed input ${X_L}= {{x_1},...,{x_n} }$ and labels ${y_L}= {{y_1},...,{y_n} }$ where ${y_L}\in (+1,-1)$, the goal is to estimate $P(y| x)$ as class membership probability for instance $x$. The probability of  $P(y| x)$ corresponds to latent function $f(x)$. Given latent function $f(x)$, the joint likelihood is stated as \cite{kuss2005assessing} (\ref{eq:6}):

\begin{equation}
P(y| f) = \prod_{i=1}^{n} P({y_i}| {f_i})
\label{eq:6}  
\end{equation}

In order to predict a new sample ${x_*}$, the distribution of the latent function related to ${x_*}$ , $f({x_*})$ is computed based on the Bayesian paradigm \cite{hernandez2016scalable}. In this study, we use RBF kernel with a length scale of one for GP classifiers.

\item Support vector machine (SVM): SVM, a statistical learning method has demonstrated good performance on a high-dimensional dataset. Primarily, SVM tries to find the hyperplane in an n-dimensional feature space that separates instances of different classes distinctively. Given linear separable input data $({x_1},{y_1}),...,({x_n},{y_n})$, ${x_i}$ as $i^{th}$ observation vector in d-dimensional feature space and ${y_i}$ as the binary class label $({y_i} \in (+1,-1))$, the hyperplane is stated as (\ref{eq:7}):

\begin{equation}
f(x) = \sum_{i=1}^{n} {w_i}.{x_i} + b 
\label{eq:7}  
\end{equation}

where $w$ is weight vector and $b$ is bias. 
\\The SVM searches for w and b to satisfy \cite{huang2018applications} (\ref{eq:8}-\ref{eq:9}):

\begin{equation}
{w_i}.{x_i} + b \geq +1 \; for \; {y_i} = +1 
\label{eq:8}  
\end{equation}

and

\begin{equation}
{w_i}.{x_i} + b \leq -1 \; for \; {y_i} = -1 
\label{eq:9}  
\end{equation}

where ${\xi_i}$ are slack variables permitting misclassification. 
The distance between these two hyperplanes is called margin, and SVM aims to maximize the margin. This problem can be stated as a quadratic optimization problem \cite{shin2005one} (\ref{eq:10}):

\begin{equation}
min \frac{1}{2}\big\|w\big\|^2 
\label{eq:10}  
\end{equation}

subject to (\ref{eq:11}):

\begin{equation}
{y_i}\big[{w_i}.{x_i}+b]\geq +1 - {\xi_i}
\label{eq:11}  
\end{equation}

Although practical data are not linearly separable, SVM benefits kernel function (\ref{eq:12}) to map data to a higher separable dimension \cite{wang2018comparative}. 

\begin{equation}
k(x, y) = \phi(x).\phi(y)
\label{eq:12}  
\end{equation}

\item SVM with radial basis function (RBF) kernel: The most widely used kernel functions by SVM are the polynomial kernel, sigmoid kernel, and radial basis function (RBF) kernel. In this study, RBF kernel with SVM is used. RBF kernel can be computed as (\ref{eq:13}):

\begin{equation}
k(x, y) = \exp \left( -\frac{\big\|x-y\big\|}{2\sigma^2} \right)
\label{eq:13}  
\end{equation}

where $\sigma$ is RBF hyper-parameter.
SVM performance is sensitive to the kernel’s parameter value \cite{wang2018comparative}.  

\item Random forest (RF): RF is an ensemble classifier, which aims to reduce the overfitting issue of decision tree algorithms by employing a parallel group of decision trees named component predictors. RF, taking advantage of the bootstrapping method, randomly generates n-sample training subset from the original training set at m times to create m independent decision trees \cite{zhang2018ice}. Moreover, each decision tree is built on a randomly selected subset of features. The final decision is made by the majority votes’ component predictors. The number of decision trees in this study is set to 10. 

\item Adaboost: Generally, boosting learning process is built on achieving a highly precise model by combining multiple inaccurate models named weak learners. Adaboost, the most well-known boosting algorithm, employs a set of base learners in an iterative process to boost the initial weak base learner to a strong learner. During this process, Adaboost seeks to minimize the error by updating the weight of instances and increasing the weight of misclassified samples over each training step \cite{subasi2018sensor}. Adaboost has demonstrated good performance against noise. Here, the number of weak classifiers is put to 50. 

\item Multi-layer perceptron neural network (MLP): MLP is a feed-forward neural network consisting of an input layer, one or several hidden layers, and an output layer. Hidden layers transfer signals from the input layer to the output layer. In MLP, adjacent layers are fully connected. MLP minimizes error through adjusting the weights of connections between the layers by employing a backpropagation algorithm \cite{zhao2012recognition}. 

\end{itemize}


\section{Discussions and results}
\label{sec:discussion_result}
Statistical reports and discussions are compiled in this section. First and foremost, we demonstrate the empirical results obtained by variant classifiers processing features extracted by pre-trained CNNs. We then examine the epistemic uncertainty using neural networks, which are the preferred models in the defect detection task.

\begin{figure}[!t]
 \begin{subfigure}{4.35cm}
     \centering
     \includegraphics[width=4cm]{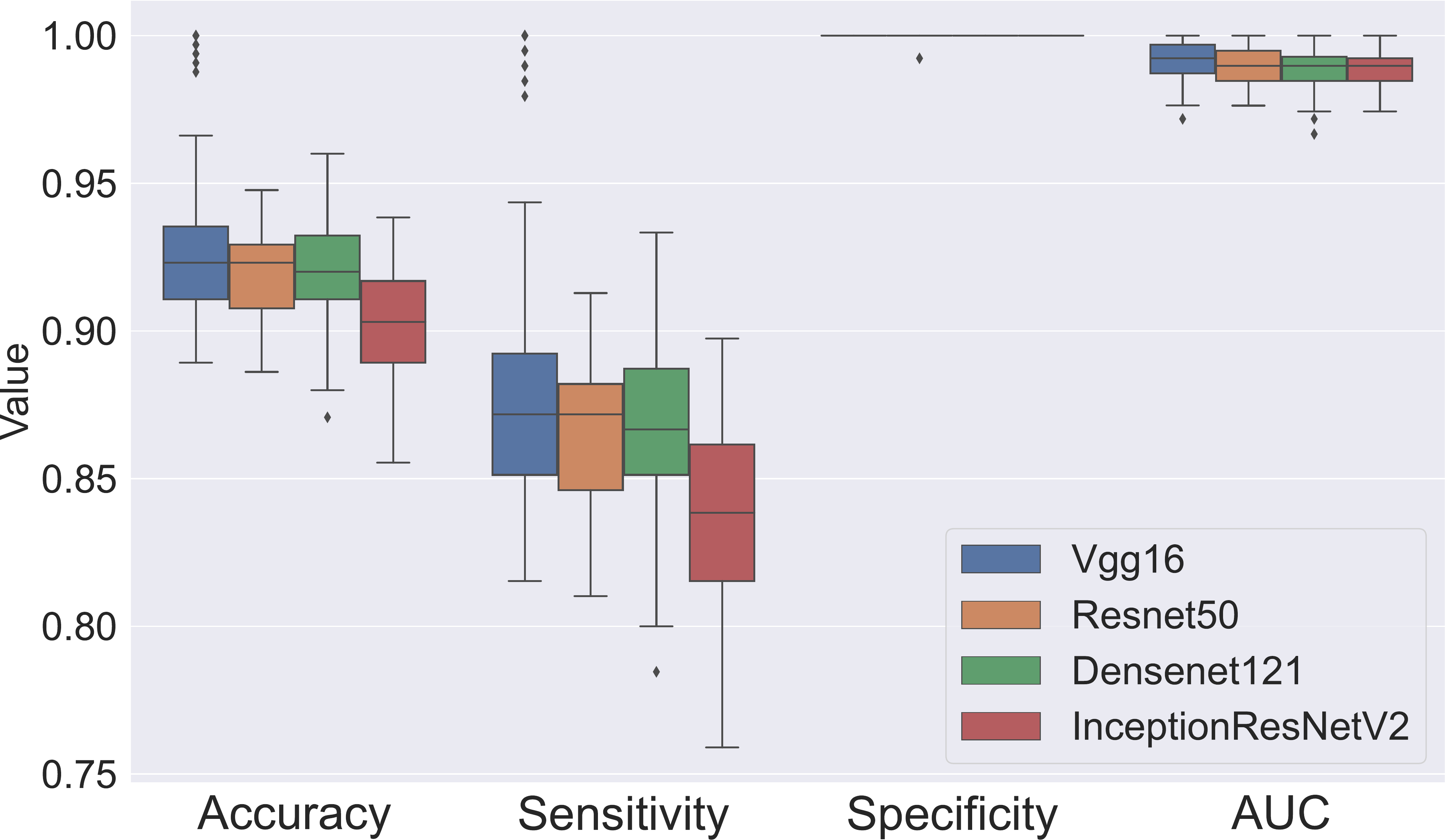}
     \caption{KNN}
   \end{subfigure}
   \begin{subfigure}{4.35cm}
     \centering
     \includegraphics[width=4cm]{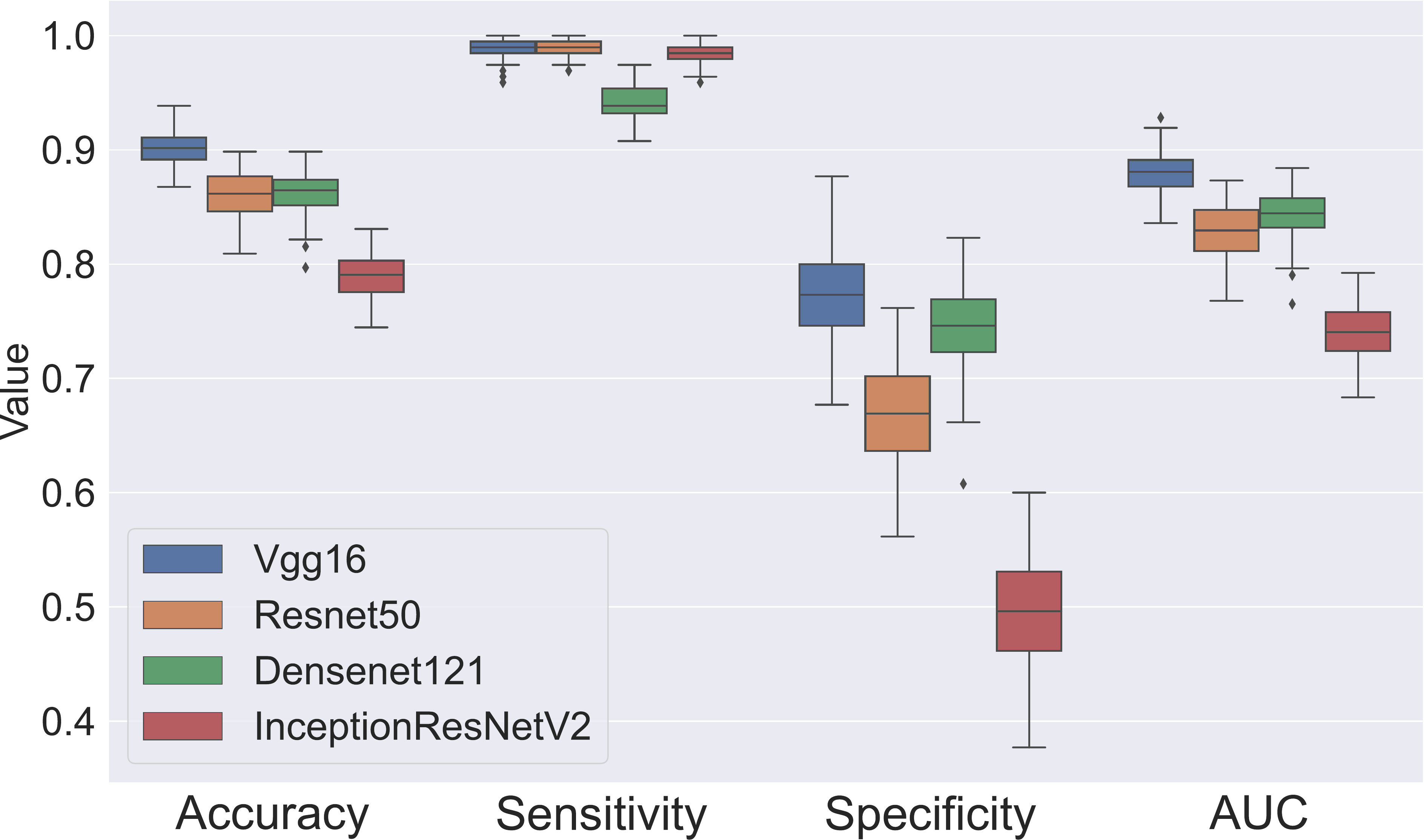}
     \caption{NB}
   \end{subfigure}
 \begin{subfigure}{4.35cm}
     \centering
     \includegraphics[width=4cm]{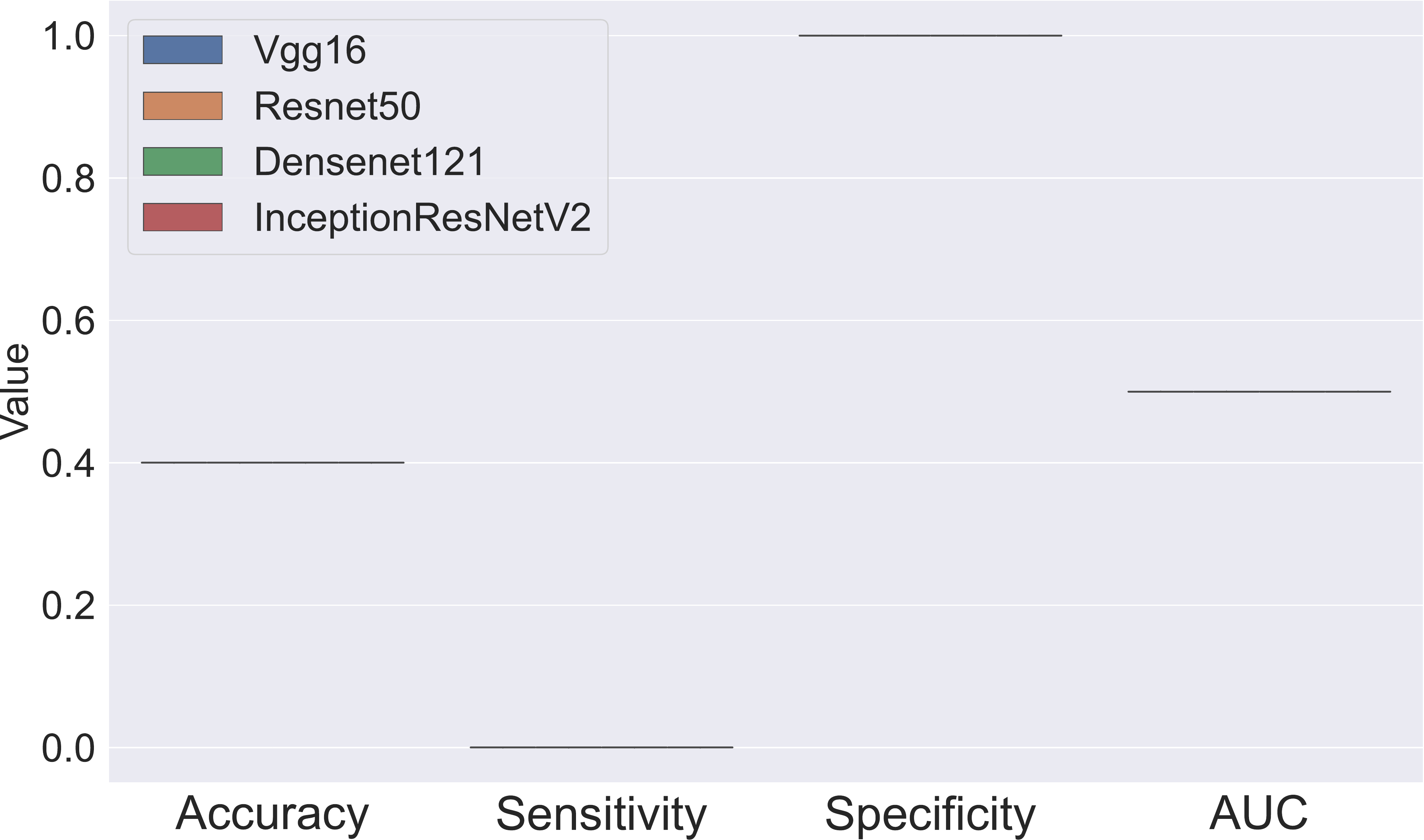}
     \caption{GP}
   \end{subfigure}
   \begin{subfigure}{4.35cm}
     \centering
     \includegraphics[width=4cm]{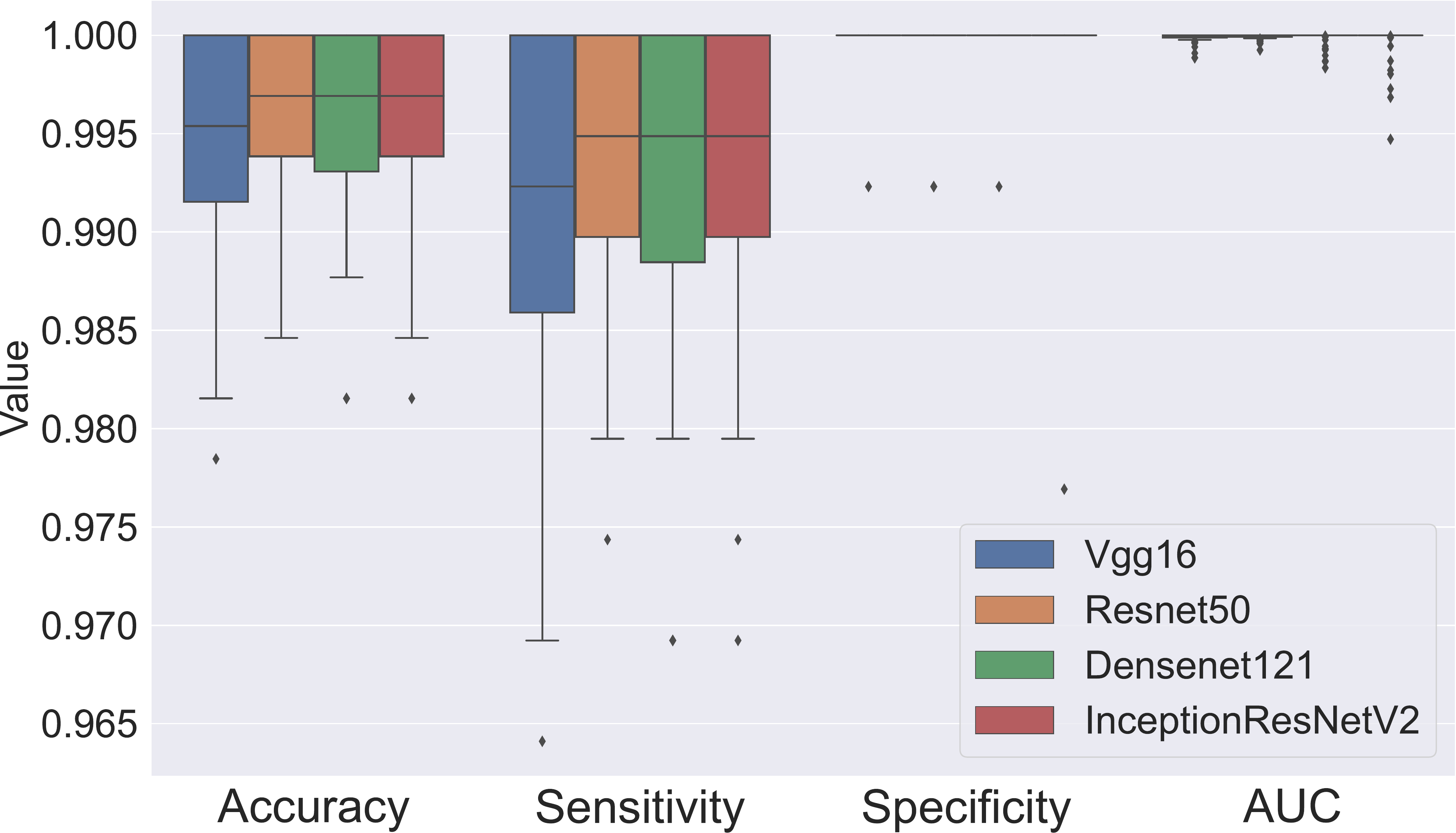}
     \caption{Linear SVM}
   \end{subfigure}
 \begin{subfigure}{4.35cm}
     \centering
     \includegraphics[width=4cm]{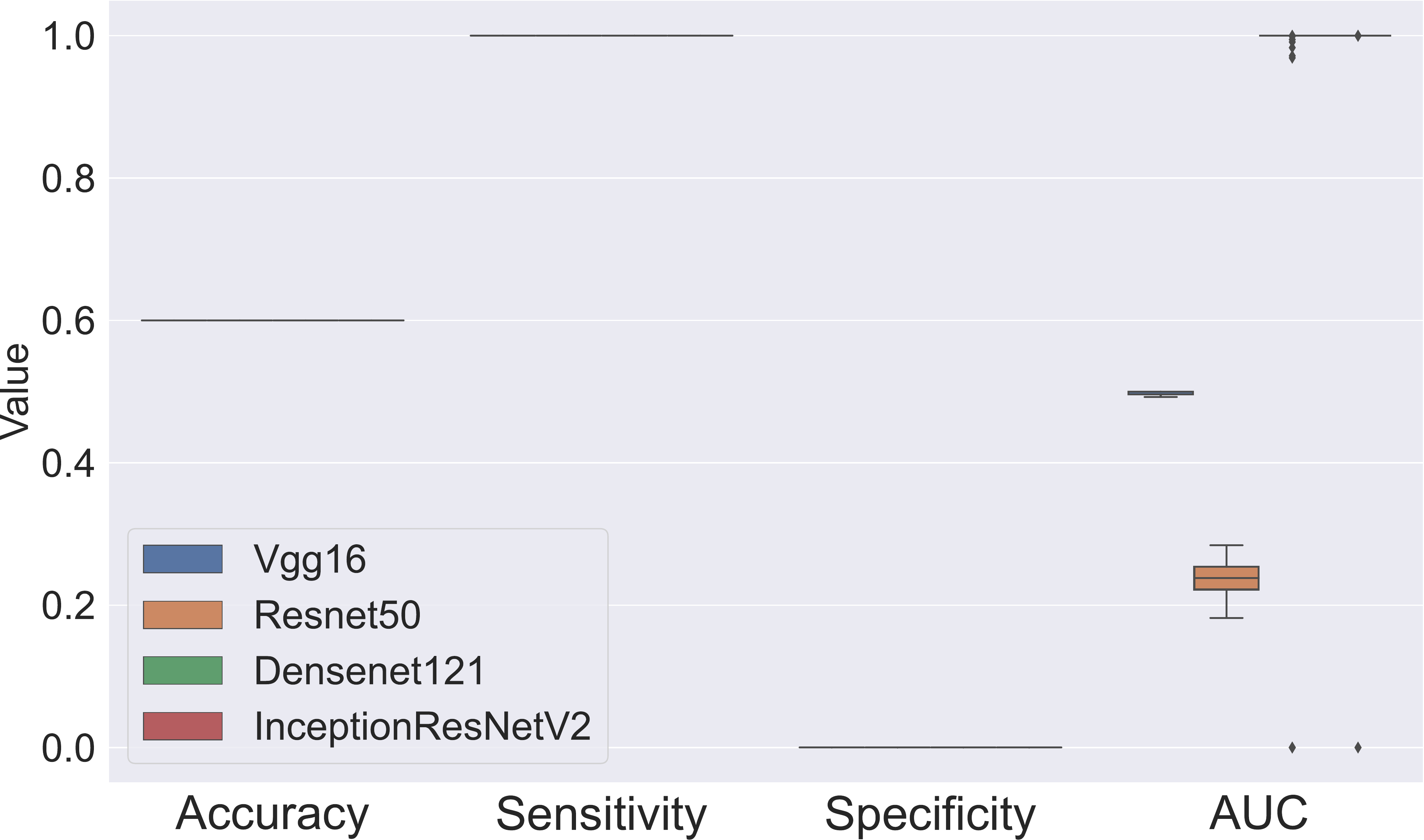}
     \caption{RBF SVM}
   \end{subfigure}
   \begin{subfigure}{4.35cm}
     \centering
     \includegraphics[width=4cm]{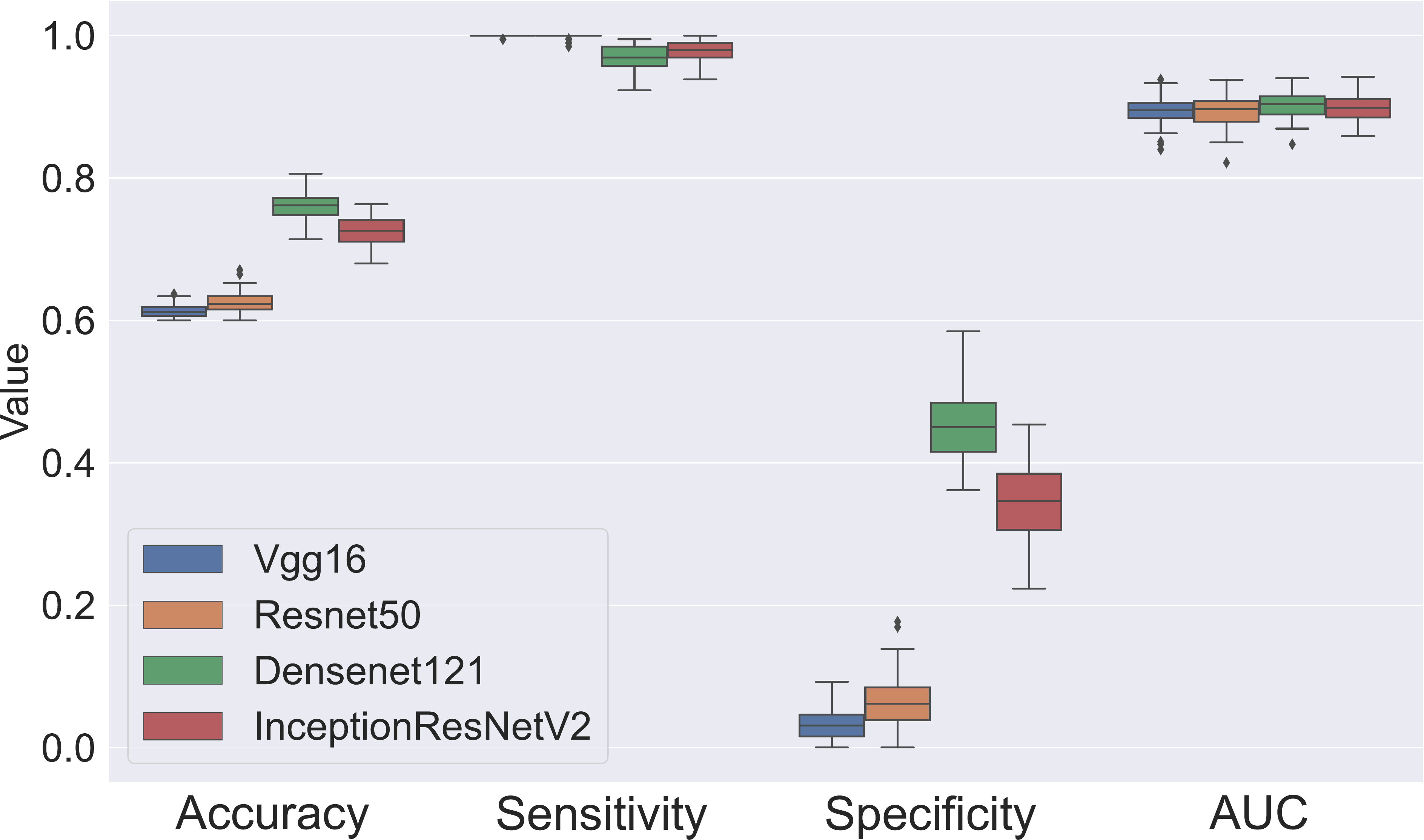}
     \caption{RF}
   \end{subfigure}
 \begin{subfigure}{4.35cm}
     \centering
     \includegraphics[width=4cm]{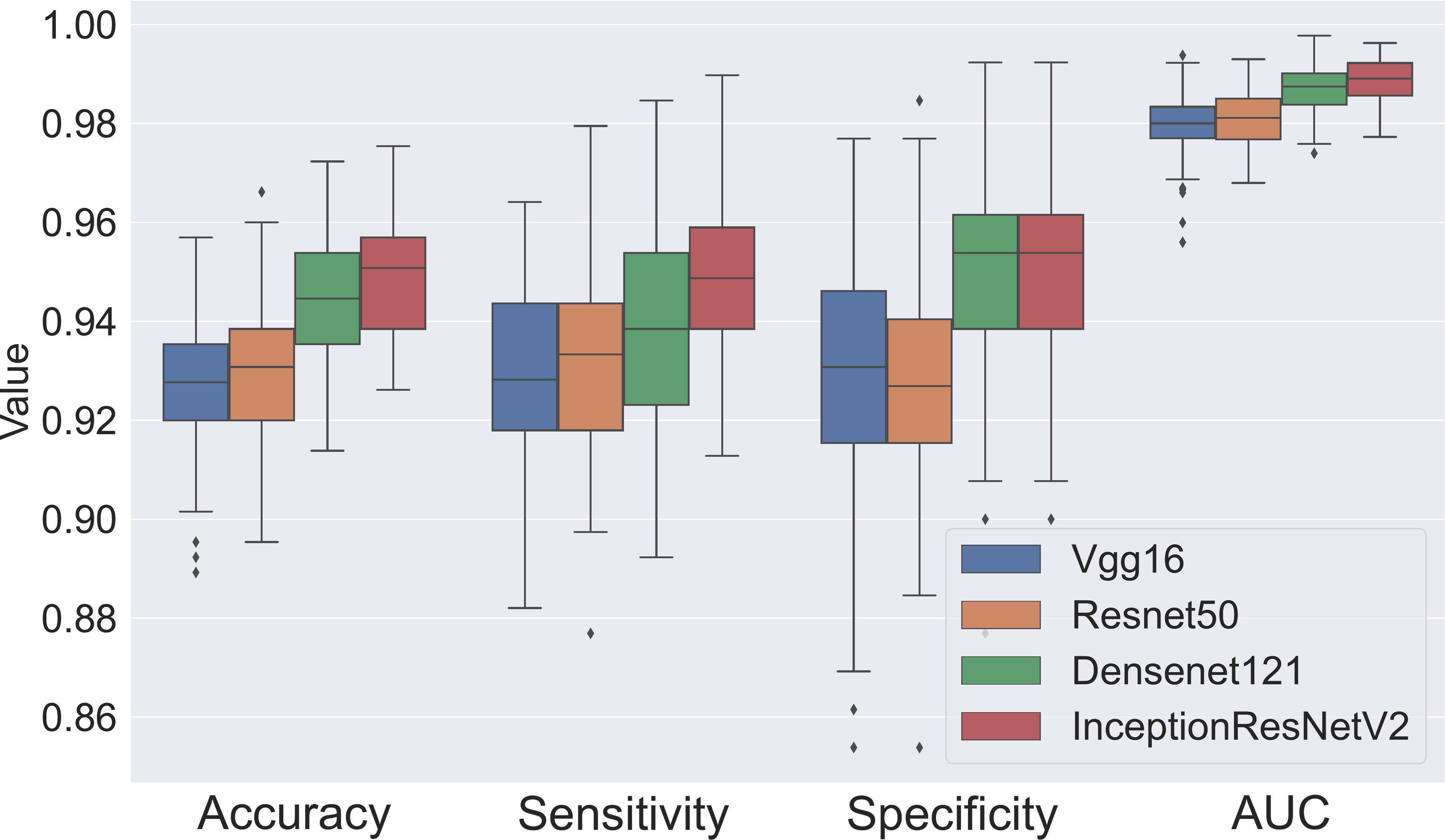}
     \caption{Adaboost}
   \end{subfigure}
   \begin{subfigure}{4.35cm}
     \centering
     \includegraphics[width=4cm]{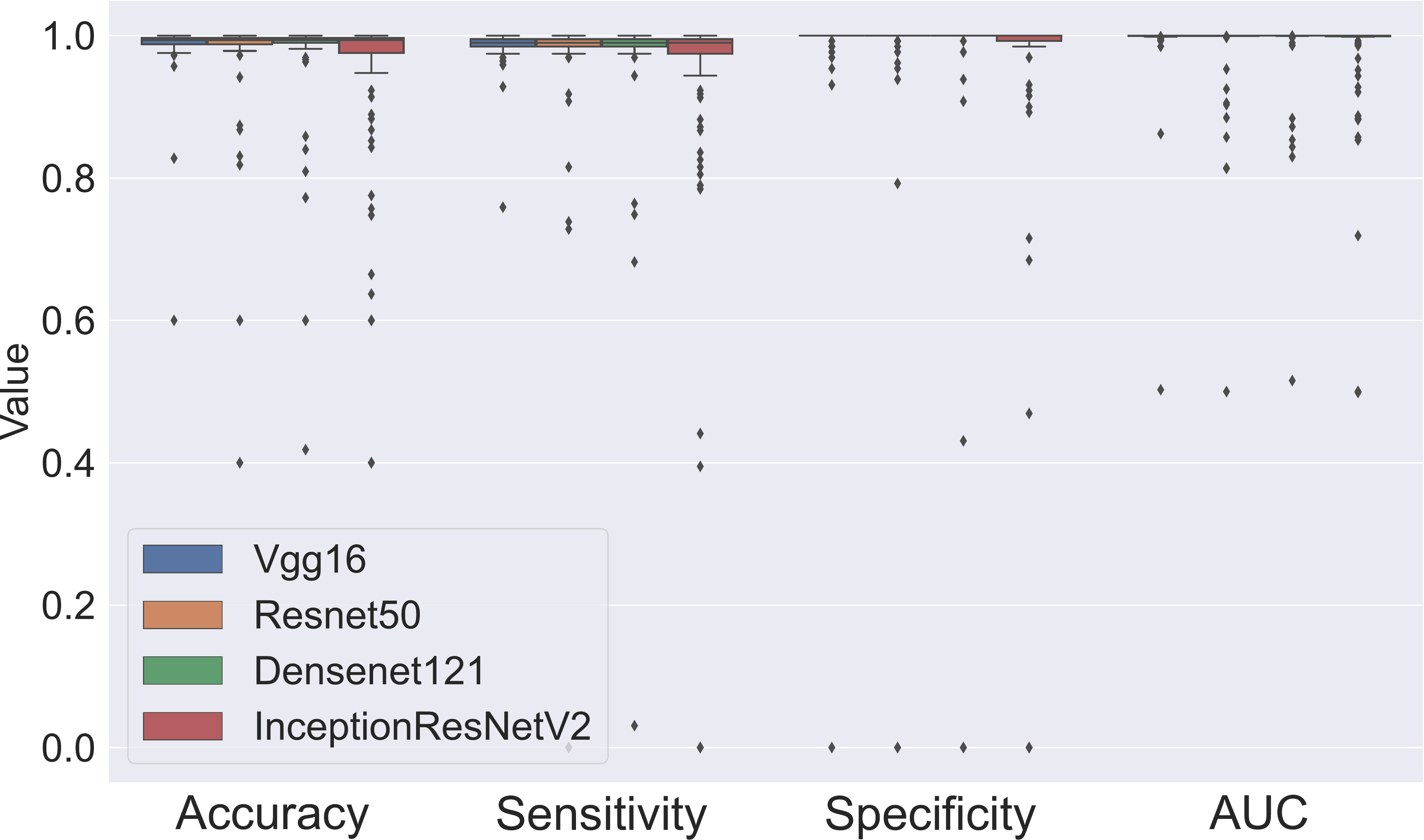}
     \caption{MLP}
   \end{subfigure}  
   \caption{Distribution of evaluation metrics of eight classifiers using four pre-trained CNNs for feature extraction. Considering four metrics together, Linear SVM and MLP indicate the best performance among the considered classifiers, while GP and RBF underperform. Note that the performance of classifiers is heavily dependent on the quality of features extracted by convolutional layers of four CNN-based models.}
   \label{fig:vgg_boxplt}
\end{figure}

Model evaluation is performed by accuracy, sensitivity, specificity, and area under the roc curve (AUC) metrics. Each classifier is trained 100 times using features obtained from pre-trained CNNs to achieve valid results. Following each run, the performance metrics are computed and then the box plots are depicted. Fig. \ref{fig:vgg_boxplt} shows the box plots for the casting product dataset, respectively for the accuracy, sensitivity, specificity, and AUC metrics. According to these plots, the performance of classifiers differs significantly based on hierarchically extracted features by convolutional layers of CNN-based models. Among the considered classifiers, linear SVM and MLP demonstrate to be the most accurate. This is because simple hyperplanes are exerted to separate features of different classes. Regrettably, the more the number of features, the more difficult it is to classify samples by RBF SVM and GP algorithms. Due to the use of the covariance matrix and its inversion, the sensitivity value for GP and the specificity value for RBF become zero, causing them to be unreliable.

To holistically compare variant architectures for feature extraction, every single classifier is trained and evaluated 100 times. The average value is then calculated from all the predicted values. All mentioned performance metrics are calculated and documented in Table \ref{tab:comparative result}. As it indicates, none of the models outperforms the other ones. The linear SVM classifier also provides the best results for each model.


\begin{table}[!t]
\centering
\small
\caption{Performance comparison between 4 architectures and 8 classifiers (32 combinations) using casting product images. All values are given in percent.}\label{tab:comparative result}
\resizebox{\columnwidth}{!}{%
\begin{tabular}{| c| c | c | c | c | c |}
\hline
Model & Classifier & Accuracy & Sensitivity & Specificity & AUC \\
\hline
{\multirow{8}{*}{\rotatebox[origin=c]{90}{VGG16}}} & Nearest Neighbors &   92.3$\pm$1.5    &   86.7$\pm$2.5    &   1.0     &    99$\pm$0.5 \\
                                        & Naive Bayes 		& 90$\pm$1.4      &    98.7$\pm$0.8    &    77.2$\pm$3.7    &    88$\pm$1.8  \\
                                        & Gaussian Process 	&   40        &    0.0         &    1.0         &    50      \\
                                        & Linear SVM 		    &   \bf99$\pm$0.5    &    99.1$\pm$0.8    &    99.9        &    99.9    \\
                                        & RBF SVM 			&   60        &    1.0         &    0.0         &    49.6$\pm$0.2\\
                                        & Random Forest 		&   61$\pm$0.9    &    99.9        &    3.2$\pm$2.3     &    89.4$\pm$1.7\\
                                        & AdaBoost 			&   92.7$\pm$1.3  &    92$\pm$1.7      &    92.8$\pm$2.5    &    97.9$\pm$0.6\\
                                        & Neural Network 		&   \bf98.6$\pm$4.2  &    98.5$\pm$2.5    &    98.7$\pm$0.1    &    99.2$\pm$5.1\\
\hline
{\multirow{8}{*}{\rotatebox[origin=c]{90}{ResNet50}}} & Nearest Neighbors & 91.9$\pm$1.4    &    86.5$\pm$2.3    &    99.9    &    98.9 $\pm$ 0.5 \\
  & Naive Bayes 		&  86$\pm$1.8    &    98.8$\pm$0.7    &    66.8$\pm$4.5    &    82.82.3 \\
  & Gaussian Process 	&  40        &    0.0         &    1.0         &    50       \\
  & Linear SVM 			&  \bf99.6$\pm$0.4  &    99.4$\pm$0.6    &    99.9$\pm$0.1    &    99.9     \\
  & RBF SVM 			&  60        &    1.0         &    0.0         &    23.7$\pm$ 2.2\\
  & Random Forest 		&  62$\pm$1.3    &    99.8$\pm$0.2    &    6.3$\pm$3.5     &    89.3$\pm$2   \\
  & AdaBoost 			&  93$\pm$1.4    &    93.1$\pm$1.7    &    92.8$\pm$2.3    &    98.1$\pm$0.6 \\
  & Neural Network 		&  \bf96.7$\pm$10.2 &    96.1$\pm$14.4   &    97.5$\pm$14.1   &    98.5$\pm$6   \\
\hline
{\multirow{8}{*}{\rotatebox[origin=c]{90}{DenseNet121}}} & Nearest Neighbors & 91.9$\pm$1.5    &    86.5$\pm$2.6    &    1.0    &    98.8$\pm$0.6 \\
  & Naive Bayes 		&  86.2$\pm$1.6  &    94.1$\pm$1.6    &    74.2$\pm$4      &    84.3$\pm$2   \\
  & Gaussian Process 	&  40        &    0.0         &    1.0         &    50       \\
  & Linear SVM			&  \bf99.5$\pm$0.4  &    99.1$\pm$0.7    &    99.9        &    99.9     \\
  & RBF SVM 			&  60        &    1.0         &    0.0         &    94.9$\pm$21.8\\
  & Random Forest 		&  76.1$\pm$1.8  &    96.8$\pm$1.5    &    45$\pm$4.8      &    90.2$\pm$1.7 \\
  & AdaBoost 			&  94.3$\pm$1.2  &    93.8$\pm$1.9    &    94.9$\pm$2      &    98.6$\pm$0.4 \\
  & Neural Network 		&  \bf97.2$\pm$8.5  &    97.2$\pm$10.5   &    97.2$\pm$15.1   &    98.7$\pm$5.7 \\
\hline
{\multirow{8}{*}{\rotatebox[origin=c]{90}{InceptionResNetV2}}} & Nearest Neighbors 	& 90.3$\pm$1.7    &    83.9$\pm$2.8    &    1.0    &    98.9$\pm$0.5 \\
  & Naive Bayes 		&  78.8$\pm$2    &    98.5$\pm$0.8    &    49.4$\pm$5.1    &    73.9$\pm$2.5 \\
  & Gaussian Process	&  40        &    0.0         &    1.0         &    50       \\
  & Linear SVM 			&  \bf99.6$\pm$0.3  &    99.4$\pm$0.6    &    99.9$\pm$0.2    &    99.9     \\
  & RBF SVM 			&  60        &    1.0         &    0.0         &    96.9$\pm$17.1\\
  & Random Forest 		&  72.5$\pm$1.9  &    97.7$\pm$1.4    &    34.8$\pm$5.8    &    89.9$\pm$1.8 \\
  & AdaBoost 			&  \bf94.9$\pm$1.2  &    94.8$\pm$1.6    &    95$\pm$2        &    98.8$\pm$0.4\\
  & Neural Network 		&  93.3$\pm$14.3 &    92$\pm$21       &    95.2$\pm$18.1   &    96.2$\pm$11.4\\ 
\hline
\end{tabular}
}
\end{table}

It is also important to consider the network size as well as the number of features, when comparing the performance of pre-trained CNNs for defect detection. As reported in Table \ref{tab:CNNs_info}, VGG16 possesses a small number of parameters and extracts the smallest set of features. In contrast, InceptionResNetV2 extracts the largest set of features. Here, when comparing the accuracy obtained by the two best classifiers in four different architectures, the VGG16 is the most effective pre-trained CNN (see Table \ref{tab:comparative result}). Discriminative neural network and linear SVM need these informative features to seek the optimal area and achieve the best result. The bigger network of InceptionResNetV2 provides the most features and less informative content. The conclusion is that bigger networks, including ResNet50 and InceptionResNetV2, do not necessarily extract more information than smaller networks.

Before commencing the UQ analysis and discussing its importance for the defect detection task, we use the Principal Component Analysis (PCA) method to map the features extracted by four pre-trained CNN models to a new space. The high-dimensional features are first obtained using method shown in Fig. \ref{fig:transferlearning} and PCA is then applied to reduce their dimensions. This is mainly done by retaining as much information as possible in a new space, which is $q$-dimensional, where $q<p$.

\begin{figure}[t]
  \centering
  \subfloat[][VGG16]{\includegraphics[width=.5\columnwidth]{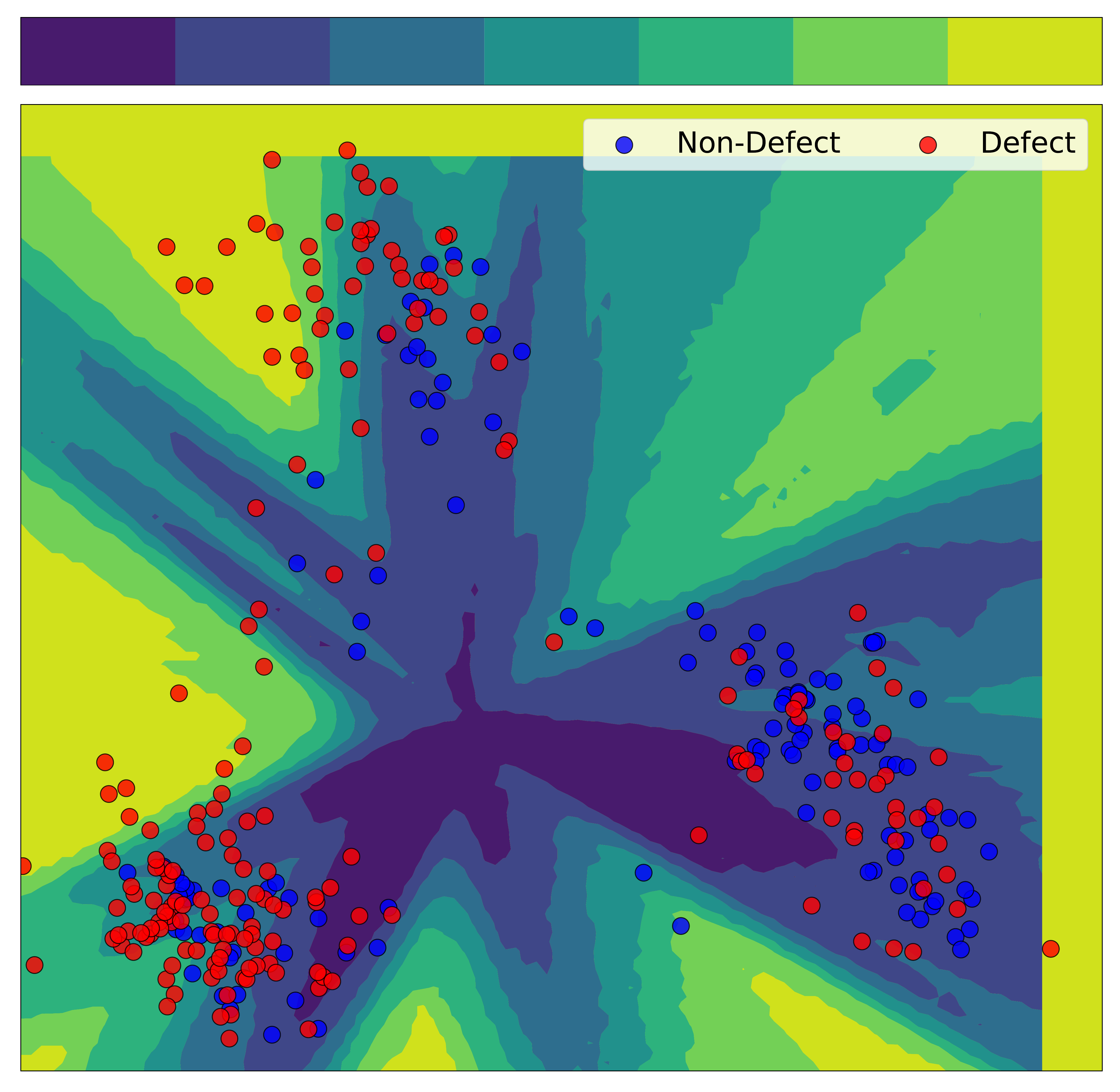}}\hfill
  \subfloat[][ResNet50]{\includegraphics[width=.5\columnwidth]{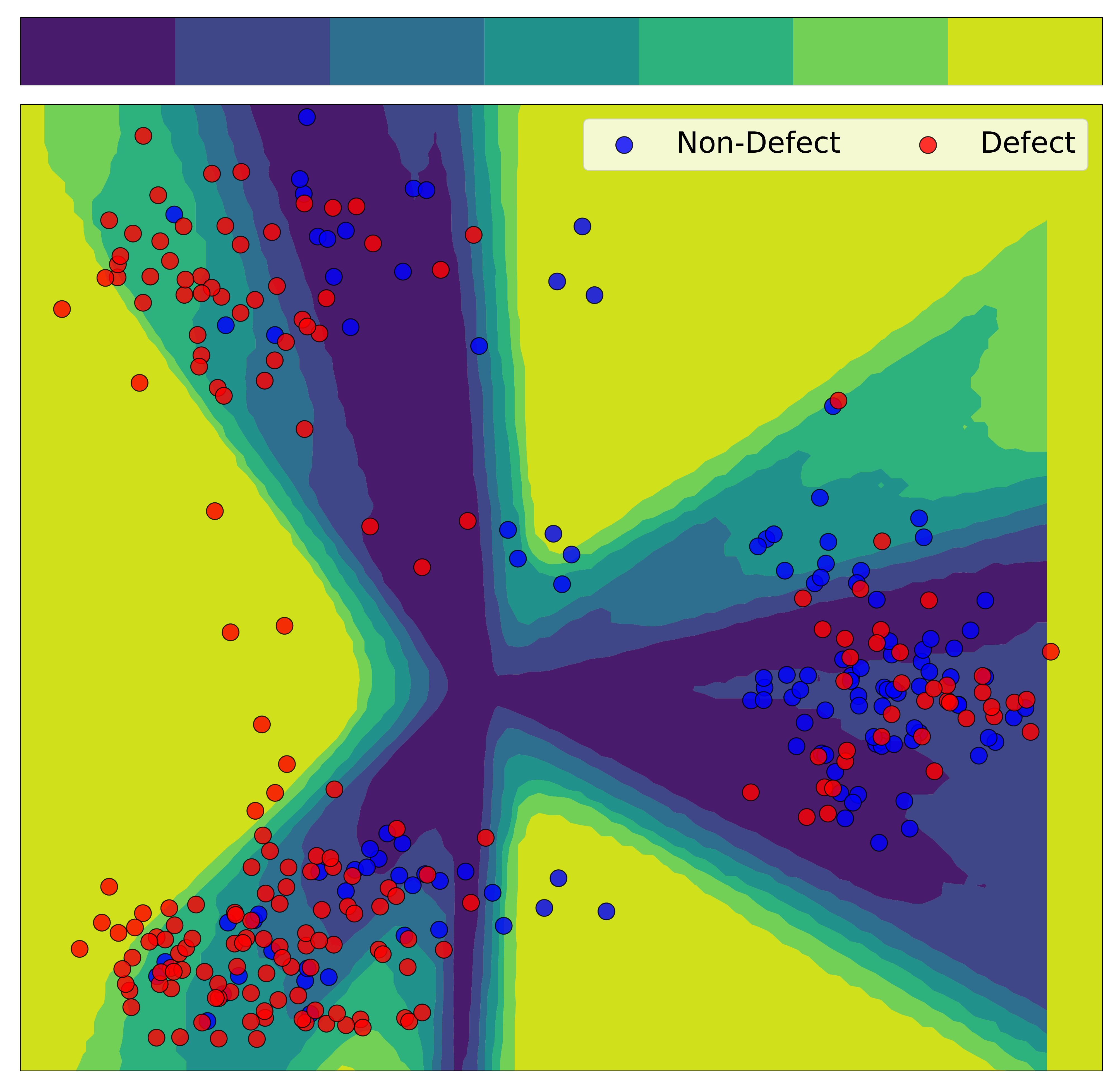}}\par
  \subfloat[][DenseNet121]{\includegraphics[width=.5\columnwidth]{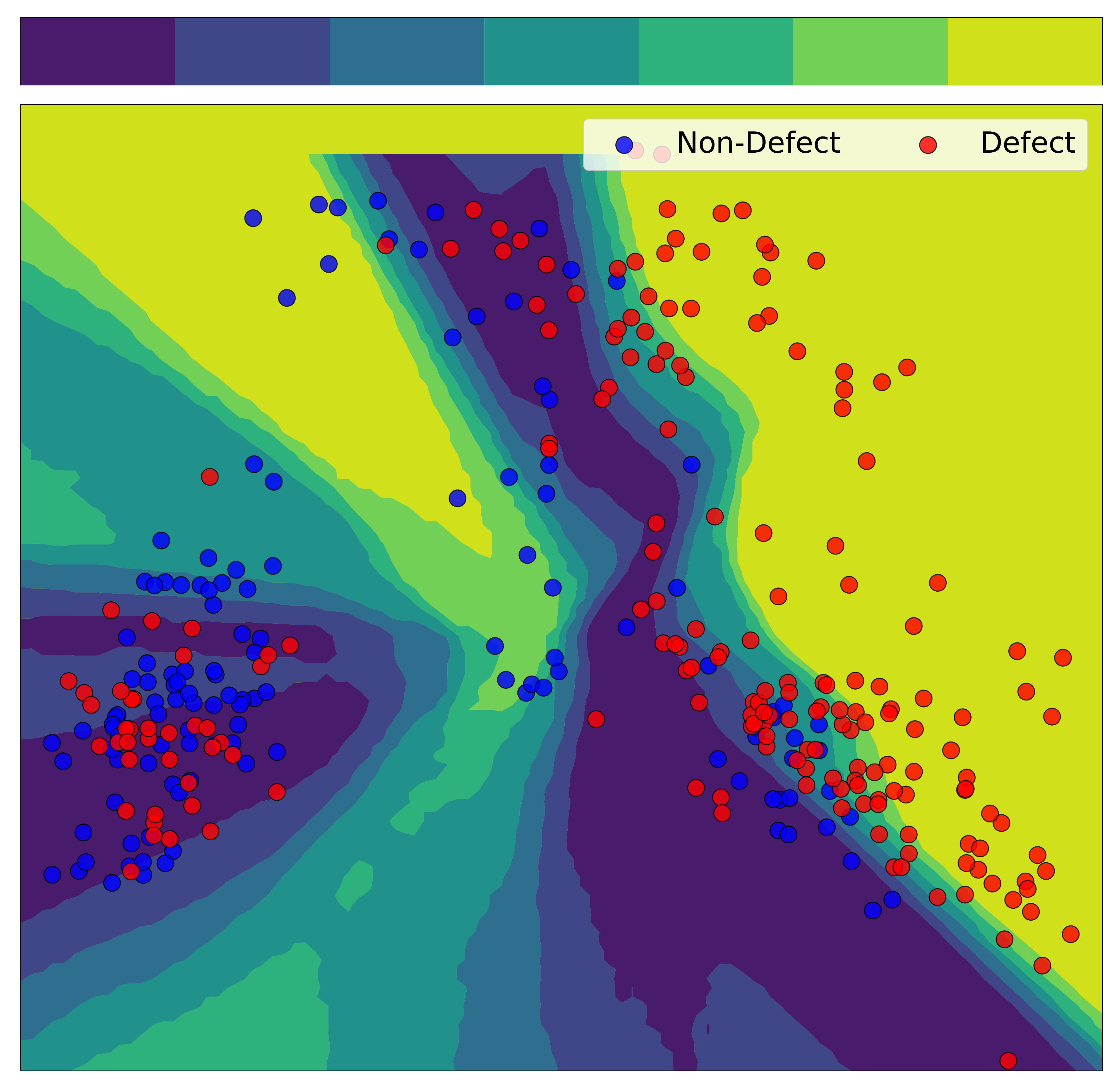}}\hfill
  \subfloat[][InceptionResNetV2]{\includegraphics[width=.5\columnwidth]{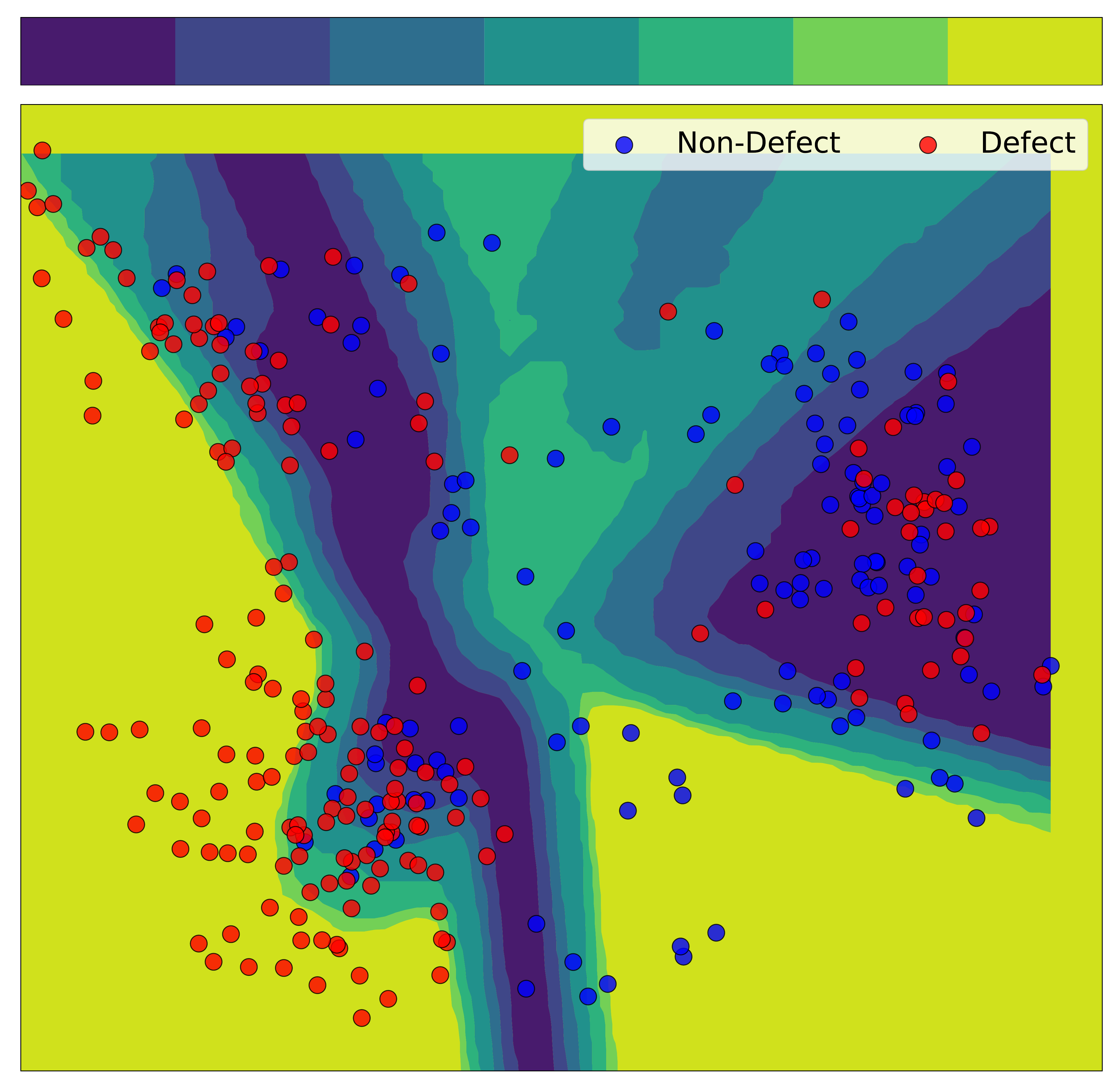}}\par
  \caption{An ensemble of Multi-Layer Perceptron models with different structures is used to quantify uncertainty. Where the color is darker, the level of uncertainty is higher. There is a high level of predictive uncertainty for samples located on dark regions of the graphs as there was no agreement among ensemble members regarding the predicted label.}
  \label{fig:Uncertainty}
\end{figure}


Ensemble of MLP models is used to examine uncertainty in the defect detection task. There are several ways to construct an ensemble of deep supervised networks. Here, the ensemble encompasses ten networks, each of which is trained independently on the entire training dataset. Explicitly, this is because more samples improve the generalization power of neural networks and learning in parallel between ensemble members is more acceptable for distribution purposes. For diversity among ensemble members, the number of layers is randomly defined between (2, 3), as well as the number of neurons within each layer is randomly selected between (256, 512), (128, 256), and (64, 128), respectively. Assuming that for a given input, an ensemble member predicts that the input belongs to the defect class with a value of x probability and the non-defect class with a value of y. If we want all ensemble members to predict that specified input, ten probabilities exist for each class. Calculating the final output probability is straightforward using (\ref{eq:1}). Now, suppose the mean probability predicts that an input belongs to the defect class and the non-defect class with 0.7 and 0.3 respectively. Using (\ref{eq:2}), the prediction entropy can be calculated as $0.7×\log(0.7)+0.3×\log(0.3)$. It is apparent that the prediction entropy becomes zero when the outcome is assigned to a class with a high probability and becomes maximum when the network does not adequate confidence about its outcome. As explained in section \ref{sec:intro}, the UQ technique computes class prediction in addition to meaningful confidence value, so the class with the bigger softmax output (argmax) for the distribution mean is considered as the predicted outcome.

Fig. \ref{fig:Uncertainty} represents the predictive uncertainty estimates for defective sample images and non-defective sample images in 2D space. This projection to 2D space is performed to ensure that the casting samples could be visualized against calculated predictive uncertainty estimates. Where the color is darker, the level of uncertainty is higher. Although the projected features in 2D space have variant locations for four pre-trained CNN models, ensemble members produce almost similar results. Regrettably, all models possess dark areas, but their extent and the number of samples within them are somewhat different. VGG16 provides the best-case scenario as there are very few samples in dark regions associated with this model. In contrast, InceptionResNetV2 is the worst since it has greater concentrations of samples in dark areas. Realistically, the uncertainty stems from a lack of training data in certain areas of the input domain. Since this is the case, there is a lack of incompatibility between ensemble members at inference time. The model is certain when the individual neural network models are consistent in decision-making. The graphs interpret the models and indicate there is a flaw in them in 2D space. 

\begin{figure}
  \centering
  \subfloat[][VGG16]{\includegraphics[width=.5\columnwidth, height=4cm]{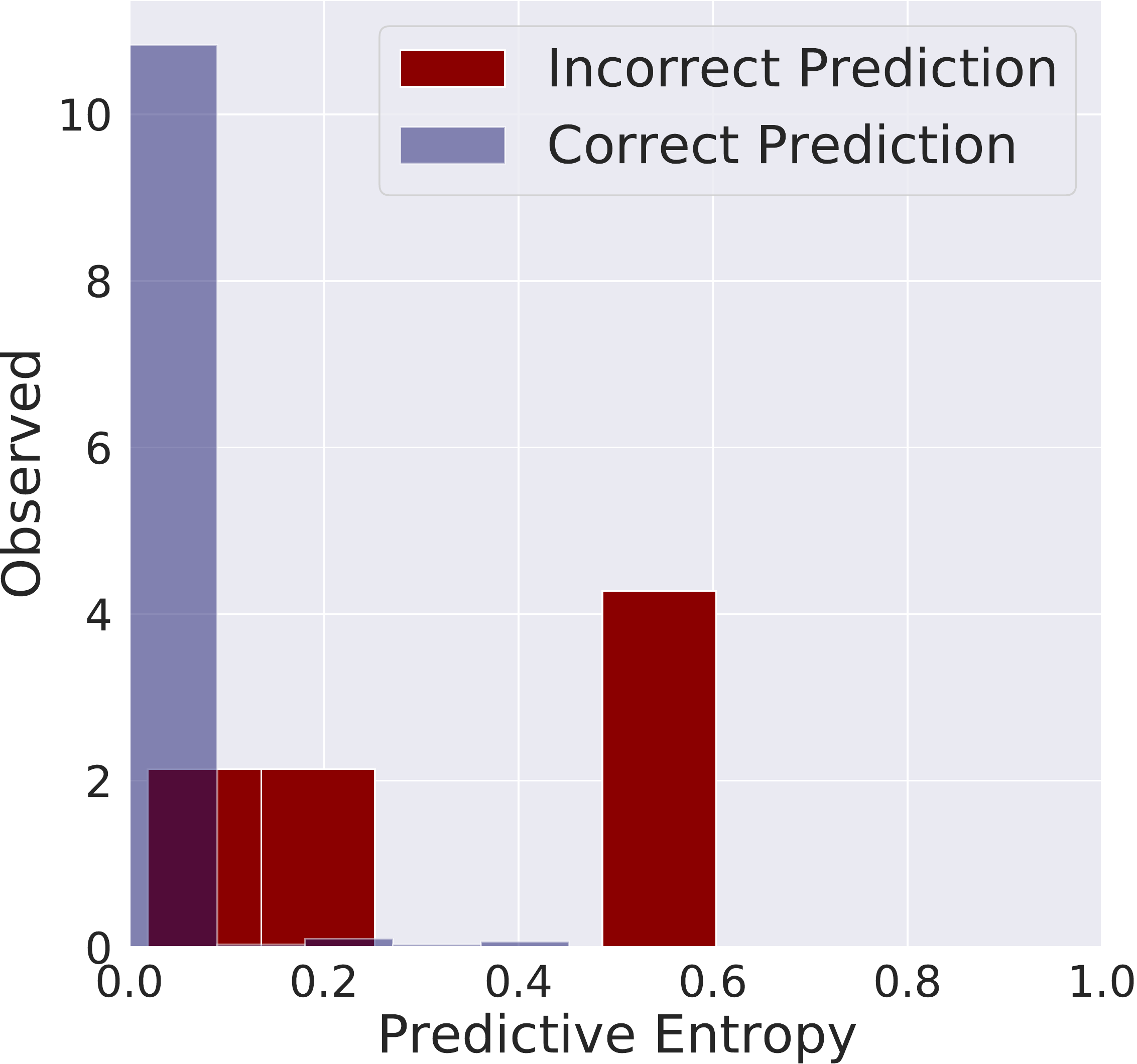}}\hfill
  \subfloat[][ResNet50]{\includegraphics[width=.5\columnwidth, height=4cm]{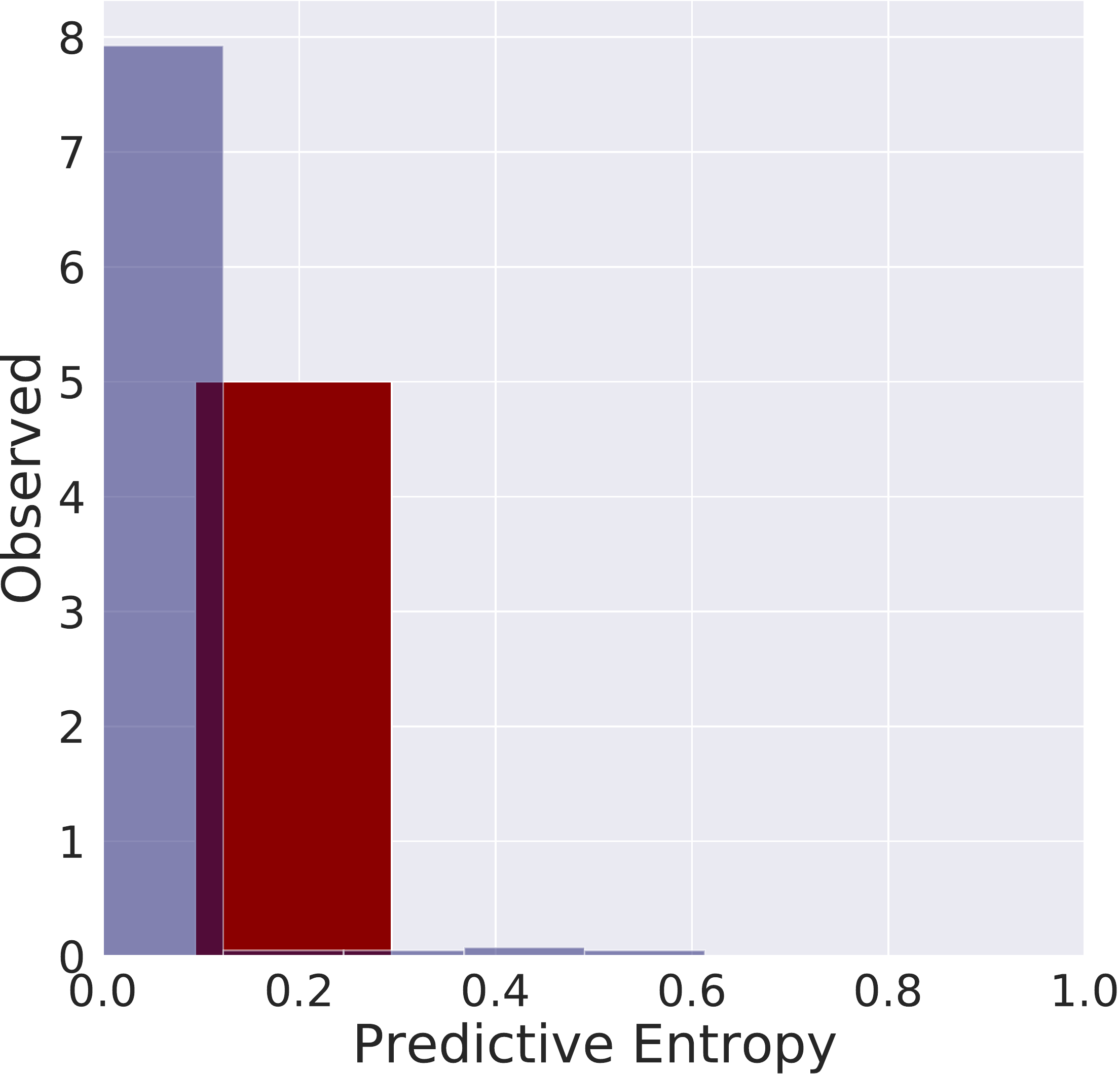}}\par
  \subfloat[][DenseNet121]{\includegraphics[width=.5\columnwidth, height=4cm]{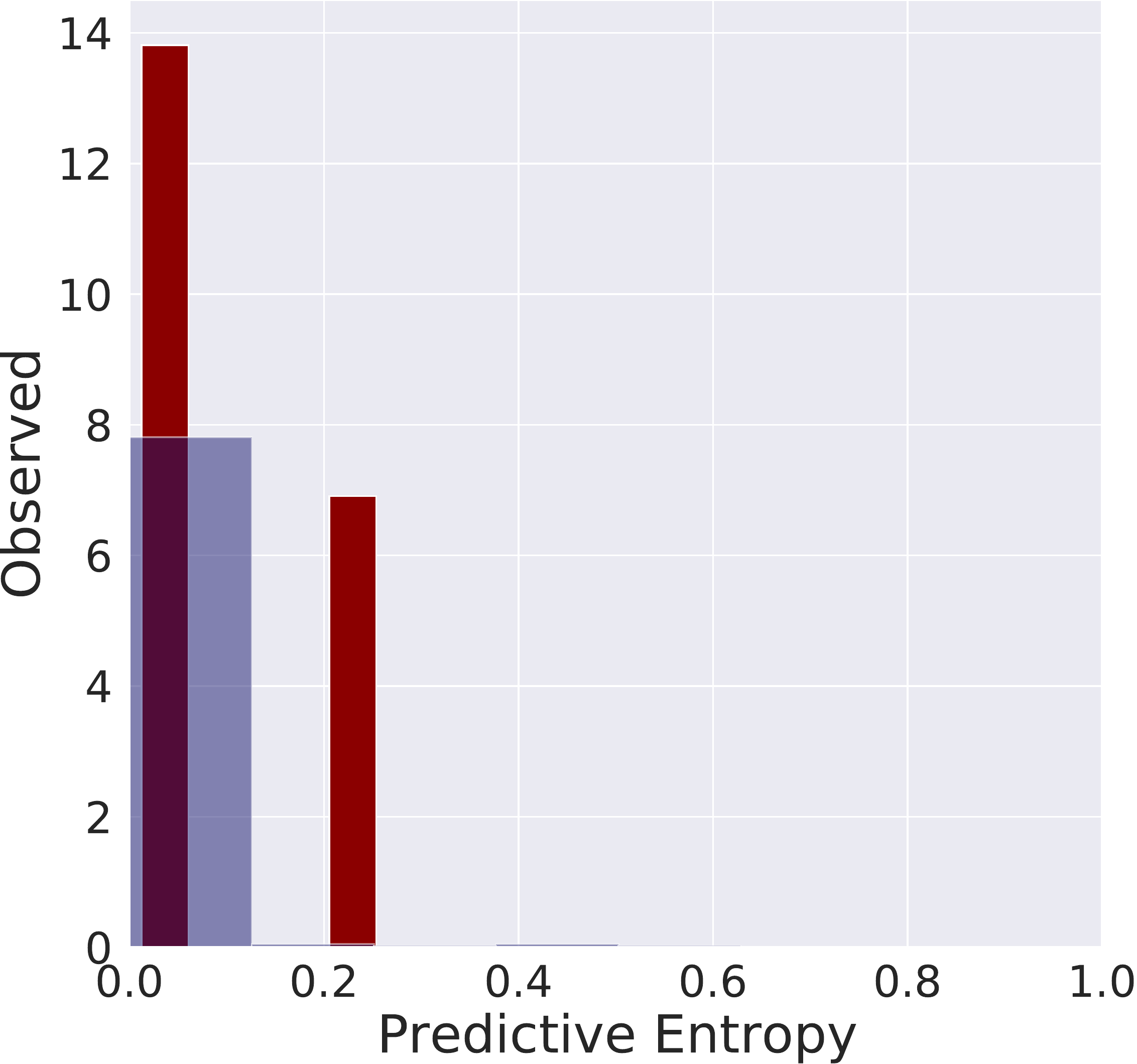}}\hfill
  \subfloat[][InceptionResNetV2]{\includegraphics[width=.5\columnwidth, height=4cm]{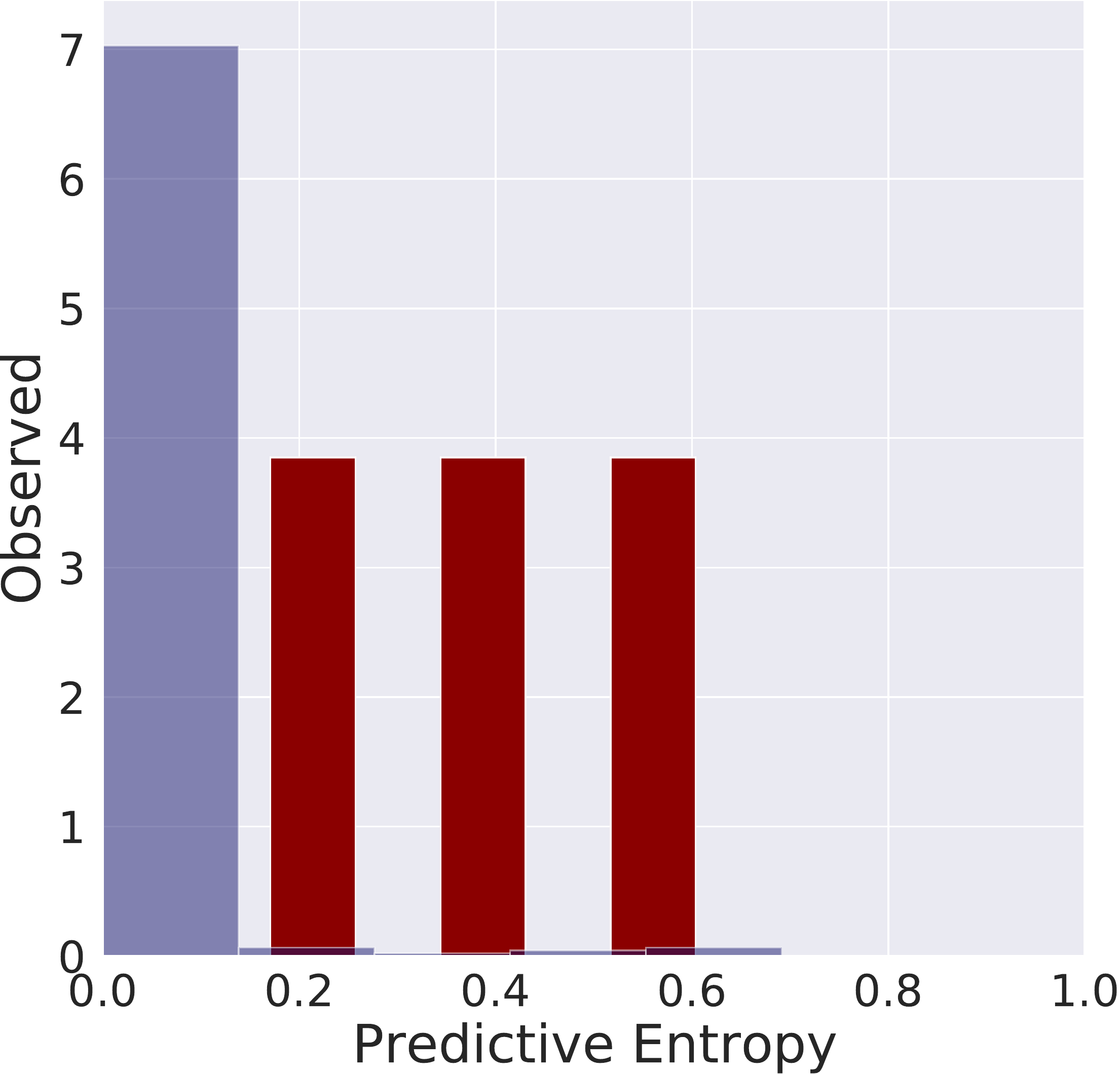}}\par
  \caption{Histogram graphs indicate the predictive entropy results for one type of UQ method (ensemble of neural networks) using features extracted from four pre-trained CNN models. In addition to uncertainty estimation, both correct classification and misclassification prediction groups are delineated in each plot. There are no notable differences between the models when it comes to best separating the two groups.}
  \label{fig:HistogramGraph}
\end{figure}

These definitions provided useful information about reducible uncertainty, but we are interested in informing more about uncertainty results obtained from UQ method considering UQ confusion matrix in Fig. \ref{fig:confmatrix}. The histogram diagrams in Fig. \ref{fig:HistogramGraph} depict the predictive entropy of the test data set for the proposed UQ technique. Four uncertainty models (ensemble of neural networks) are developed using features extracted from four pre-trained CNN models without applying PCA (all features passed to networks). In Fig. \ref{fig:HistogramGraph}, in addition to the predictive entropy is computed using (\ref{eq:2}), the class with the bigger softmax output is compared with ground truth labels, and the final result is then categorized into the incorrect classification and correct classification groups. Correct and certain predictions (TC) shift to the left of the histogram (blue peaks with entropy close to zero), while the incorrect and uncertain predictions (TU) shift from left to the right of the histogram (red peaks with entropy close to maximum). At inference time, the UQ method is able to detect that misclassification items are far from the correctly classified items, which manifest higher confidence in networks. However, there are few predictions between both groups (TC and TU) as incorrect and certain predictions (FC), correct and uncertain predictions (FU), which were erroneously flagged due to the model's inability to communicate with its confidence ideally. It seems highly likely that most of the incorrect predictions have shown along with uncertainty. Meaningfully, the model has announced,\textit{"Do not trust my uncertain predictions"}. Indeed, the image data that were misclassified, say that the model is not sure of these answers, so this is a positive development. This opens up an opportunity for engineers to examine necessarily and prevent potential losses.

Defining these terms as certainty and uncertainty is based on a boundary between them. For instance, by assuming that an entropy threshold of 0.3 is considered for the InceptionResNetV2 graph. It is observed that there are samples as TC and FC before this determined threshold. Now, if we want to change the entropy threshold to 0.5, the number of TCs and FCs increases. Accordingly, as the decision boundary changes, the number of TC, TU, FC, and FU changes as well. Considering the best-case scenario, the misclassification items with higher uncertainty (entropy close to one) and the correct classification items with higher certainty (entropy close to zero) have,  the UQ model is more conscious of its strengths and weaknesses. The farther the two blue (TC) and red (TU) couriers are, the better. The smoother or flatter the graph between the red and blue peaks shows the less confusing the model has. 

Fig. \ref{fig:HistogramGraph} provides qualitative quantification of uncertainty to illustrate a comparison between four UQ models with different architectures and is not necessarily an accurate representation of UQ estimation. As such, to calculate the uncertainty accuracy metric (\ref{eq:3}), an entropy threshold is selected between 0.1 and 0.9 as a decision boundary. Here the threshold rate of 0.4 is preferably considered in the task of defect detection. This performance metric for all models is calculated and documented in Table \ref{tab:ComparisonUncertaintyAccuracy }. The proposed UQ method based on the VGG16 outperforms the other ones to capture uncertainty. This corroborates that VGG16 is the most efficient pre-trained CNN in the defect detection of casting images.

\begin{table}[]
\caption{Comparison of uncertainty accuracy performance metric for UQ method using features extracted of four pre-trained CNN models (threshold set to 0.4).}
\label{tab:ComparisonUncertaintyAccuracy }
\resizebox{\columnwidth}{!}{%
\begin{tabular}{cc}
\hline
The architecture of the UQ method & Uncertainty Accuracy \\
\hline
VGG16 & 99\% \\
ResNet50 & 98.4\% \\
DenseNet121 &  98.7\% \\
InceptionResNetV2 & 97\% \\
\hline
\end{tabular}%
}
\end{table}

\section{Conclusion}
\label{sec:conclusion}
This work proposes the deep transfer learning method as one of the most powerful paradigms in ML for the defect detection task using the casting product images. For developing deep neural networks from scratch, thousands or even millions of images are necessary. The lack of availability large amount of casting sample images motivated us to use this procedure. Leveraging the transfer learning framework, we exploited four pre-trained CNN models including VGG16, ResNet50, DenseNet121, and InceptionResNetV2 to extract deep features from the casting sample images. During the transfer learning process, network weights are kept frozen. Extracted features were then processed by several supervised learning models. After obtaining the empirical results, the conclusion is that the MLP model and linear SVM outperform other learning models in terms of defect detection accuracy for the casting sample images. Additionally, we applied an UQ method (ensemble of MLP models) using features extracted from four pre-trained CNNs to predict the meaningful uncertainty associated with the predictions. We also used the UQ confusion matrix to evaluate the predictive uncertainty estimates. An ensemble of neural networks based on the VGG16 outperforms the other ensembles to capture uncertainty. This corroborates that VGG16 is the most efficient pre-trained CNN in the defect detection of casting images. \\ There are several possible directions for future work. It would be interesting to combine features obtained from different transfer learning algorithms to develop hybrid models. We focused on training independent networks with randomized neurons in their layers to construct four ensembles of networks. De-correlating networks 'predictions could contribute to ensemble diversity and improve performance in capturing uncertainties. Optimizing the ensemble weights can be another option to improve performance.



\end{document}